\documentclass[runningheads,a4paper]{llncs}

\usepackage{graphicx}
\usepackage{amsmath,tabu}
\usepackage{amssymb}
\usepackage[dvipsnames]{xcolor}
\usepackage{ntheorem}
\usepackage{subfigure} 
\usepackage{times}

\usepackage{url}
\newcolumntype{C}[1]{>{\centering\let\newline\\\arraybackslash\hspace{0pt}}m{#1}}

\begin{document}
\mainmatter  

\title{Error Asymmetry in Causal and Anticausal Regression}

%
%
\author{Patrick Bl{\"o}baum, Takashi Washio, Shohei Shimizu}
\authorrunning{Error Asymmetry in Causal and Anticausal Regression}

\institute{The Institute of Scientific and Industrial Research, Osaka University}

%
%

\toctitle{Error Asymmetry in Causal and Anticausal Regression}
\maketitle
\begin{abstract}
It is generally difficult to make any statements about the expected prediction error in an univariate setting without further knowledge about how the data were generated. Recent work showed that knowledge about the real underlying causal structure of a data generation process has implications for various machine learning settings. Assuming an additive noise and an independence between data generating mechanism and its input, we draw a novel connection between the intrinsic causal relationship of two variables and the expected prediction error. We formulate the theorem that the expected error of the true data generating function as prediction model is generally smaller when the effect is predicted from its cause and, on the contrary, greater when the cause is predicted from its effect. The theorem implies an asymmetry in the error depending on the prediction direction. This is further corroborated with empirical evaluations in artificial and real-world data sets.
\end{abstract}

\section{Introduction}
\label{introduction}
The nature of a prediction problem can be quite complex. Many factors lead to the observed data, such as the underlying distributions or causal relationships between variables. A general statement about the expected error for a given problem can therefore be a tough task. Especially in the domain of health care, a precise estimator is indispensable to ensure the right conclusions for the treatment of a patient. In order to make any statement about the expected error, a typical approach is to further analyze the data properties or the properties of the utilized algorithm in detail. For example, facing a noisy classification problem where there is 70\% chance that a sample belongs to class one or a 30\% chance to class two, the best guess for an unknown sample would always be class one, therefore there exists a 30\% risk of a misclassification. An analog statement is possible for regression problems, where there exists an infimum over the error due to some noise affecting the data. In this paper, we show that this infimum depends on the prediction direction. We further emphasize that the optimal estimator w.r.t. to a risk minimization may not necessarily minimize the prediction error. Therefore, detailed knowledge about the underlying problem can give additional useful information not only in the area of machine learning but also with regards to general prediction tasks.

We give a novel insight regarding the intrinsic causal structure of a problem and the resulting implications for the prediction error under the assumptions of an additive noise and an independence between mechanism and cause. Our work is based on the ideas by \cite{OnCausal}, who claimed that, under certain assumptions, the underlying causal directions of variables have important implications for various machine learning scenarios. Similar as in \cite{OnCausal}, we address the setting of two observable variables where one variable is the cause and the other variable is the effect. We analyze the expected prediction error of regression tasks when the true data generating function is utilized as prediction model and draw attention to the fact that the key factors for the expected error are fundamentally different if the effect is predicted by the cause and if the cause is predicted by the effect. While the error generally depends heavily on the noise, the actual shape of the underlying function that generated the data is another crucial key factor that differs between the prediction directions. This has not been recognized in the past and is explicitly pointed out in this paper. In particular, we formulate a fundamental theorem which states that an asymmetry of the prediction error in regard to the prediction direction between two variables exists.

\section{Background and Problem Setting}
\label{sec:background}
In the following, we provide a short overview of the background theories and give a simple motivating example.

\subsection{General causal structure and notation}
Graphical models provide a framework to describe the causal structure of a set of variables that represent their joint distribution \cite{Pearl:2009:CMR:1642718}. These variables are vertices in a directed acyclic graph, where a direct causal influence of a variable $X_i$ on a variable $X_j$ is indicated by an arrow between these two vertices.

For the theoretical analysis, we consider a two variable model with an observable \emph{cause} variable $C$, an observable \emph{effect} variable $E$ and a latent noise variable $N_E$ with $\mathbb{E}[N_E] = 0$. The noise $N_E$ only affects the effect $E$ and is assumed to be independent of $C$. We denote the probability distribution of a random variable $X$ by $P(X)$ and the corresponding density by $p(X)$. For simplicity, $p(X)$ denotes either the density distribution or a density value with respect to $X$ depending on the context. This notation is also used for the joint and conditional distributions. All densities are assumed to be strictly positive $p(X) > 0$.

\begin{figure}[t]
\hspace*{0.01\columnwidth}\subfigure[]{
  \centering
  \includegraphics[width=0.31\columnwidth]{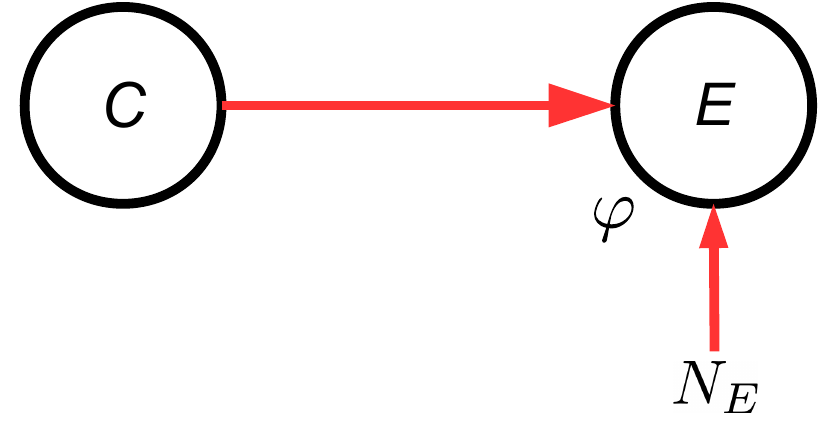}
  \label{fig:GeneralCausalStructure}
}
\subfigure[]{
  \centering
  \includegraphics[width=0.31\columnwidth]{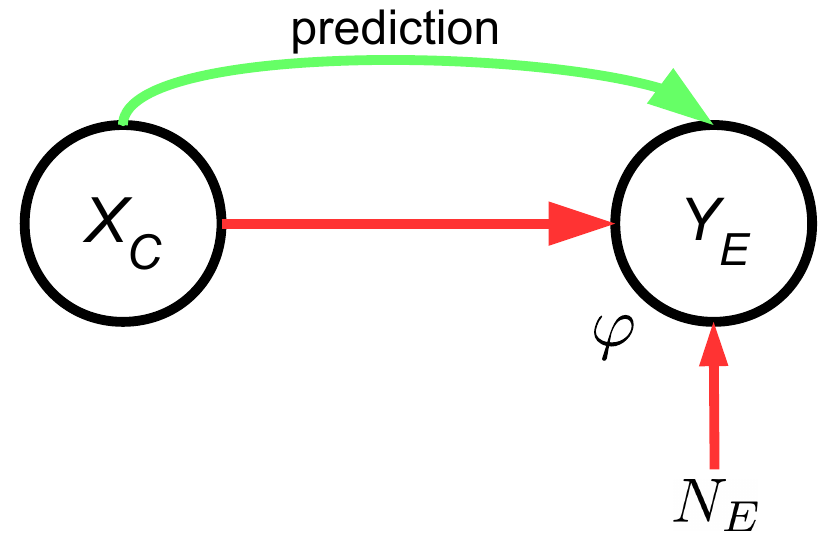}
  \label{fig:CausalPrediction}
}
\subfigure[]{
  \centering
  \includegraphics[width=0.31\columnwidth]{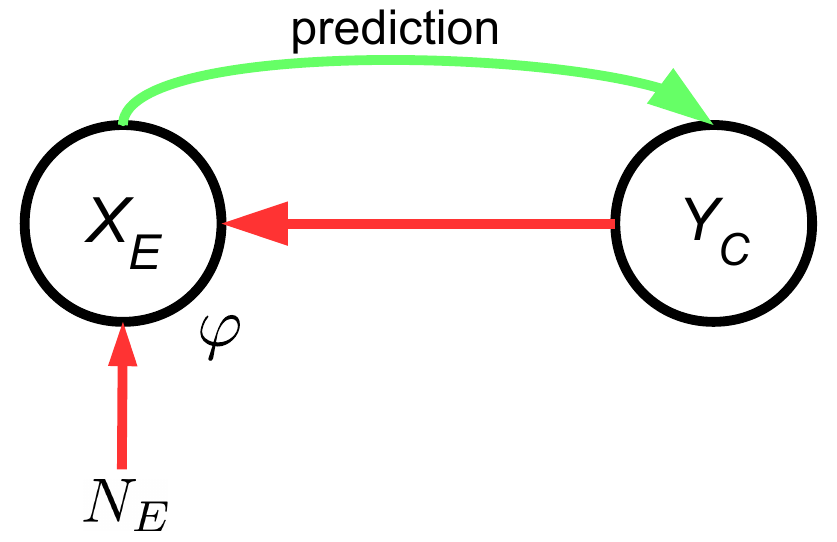}
  \label{fig:AnticausalPrediction}
}
\caption{\textbf{(a)} An illustration of a simple causal structure with two variables; a cause $C$ and an effect $E$. The effect is influenced by some noise $N_E$ that is assumed to be independent of $C$. Between cause and effect exists a mechanism $\varphi(C, N_E) = E$ which determines the effect. \textbf{(b)} The effect $E$ is predicted from the cause $C$. \textbf{(c)} The cause $C$ is predicted from the effect $E$.}
\end{figure}

The causal relationship between cause and effect is defined by a \emph{mechanism} $E = \varphi(C, N_E)$, where $\varphi$ determines the effect $E$ given cause $C$ and noise $N_E$ as illustrated in Figure \ref{fig:GeneralCausalStructure}. We assume that observed data were generated following this intrinsic causal relationship. In the following, we use the term ``mechanism'' for $\varphi$ and the conditional $p(E|C)$.\footnote{The conditional $p(E|C)$ is defined by $E = \varphi(C, N_E)$ and $p(N_E)$.}

\subsection{Problem setting} 
In general prediction problems, the goal is to predict a target variable $Y$ from a predictor variable $X$. Prediction models typically aim to learn the mapping from $X$ to $Y$ based on observed data that is sampled from the joint distribution $p(X, Y)$. Further knowledge about the causal relationship is rarely available and requires additional domain knowledge. If this is not available, the causal direction can be inferred by \emph{causal inference} methods that try to determine whether $X$ or $Y$ is cause or effect, respectively. Knowledge about the causal direction is particularly important in the domain of biomedicine, where it e.g. provides a better understanding of the relationship between symptoms and diseases. For our analysis, the causal relationship is assumed to be known and we rather focus on the implications for the prediction.

Following the ideas of \cite{OnCausal}, we differ between predicting the effect from the cause $C \rightarrow E$ (causal prediction) and predicting the cause from the effect $E \rightarrow C$ (anticausal prediction) as shown in Figure \ref{fig:CausalPrediction} and \subref{fig:AnticausalPrediction}, respectively.

\textbf{Example (Rain problem):} A simple example for a causal problem would be to predict if a street is wet $E = 1$ or dry $E = 0$ based on the observation whether it is raining $C = 1$ or not $C = 0$. In this problem, it is clear that the street becomes wet as soon as it starts to rain. Therefore, there is a clear causal relationship where the rain $C$ is the cause and the wet street $E$ is the effect. It is unlikely to observe a wet street without rain, but due to other occurrences, it may happen that a wet street is observed even when it is not raining. This can be seen as some noise $N_E$ influencing the effect. For simplicity, it is assumed that the street instantly dries out as soon as the rain stopped, so there is no dependency between $C$ and $N_E$. The causal conditional $P(E|C)$ is given in Table \eqref{eq:causalMechanism} and the anticausal conditional $P(C|E)$ in Table \eqref{eq:anticausalMechanism}. The probability of observing a wet street without any rain is therefore $P(E = 1|C = 0) = N_E$. The conditional $P(C|E)$ represents the rain problem from an anticausal perspective where the task would be to predict if it is raining based on whether the street is wet or dry. Note that $P(C)$ does not occur in $P(E|C)$ in Table \eqref{eq:causalMechanism}, but $P(C|E)$ heavily depends on it.

{\renewcommand{\arraystretch}{1.4}
\begin{table}
\caption{The causal conditional $P(E|C)$ and the anticausal conditional $P(C|E)$ of the rain problem. Here, the noise is expressed as $N_E \in [0, 1]$.}
\begin{subequations}
\begin{equation}
P(E|C) = \begin{tabu}{c|c|c}
 & C = 0 & C = 1 \\ 
\hline 
E = 0 & 1 - N_E  & 0 \\ 
\hline 
E = 1 & N_E & 1 \\ 
\end{tabu}
\label{eq:causalMechanism}
\end{equation}
\begin{align}
P(C|E) &= \nonumber \\
&\begin{tabu}{c|c|c}
 & E = 0 & E = 1 \\ 
\hline 
C = 0 & 1 & \frac{N_E P(C = 0)}{N_E P(C = 0) + P(C = 1)} \\ 
\hline 
C = 1 & 0 & \frac{P(C = 1)}{N_E P(C = 0) + P(C = 1)} \\ 
\end{tabu},
\label{eq:anticausalMechanism}
\end{align}
\end{subequations}
\end{table}

\textbf{Error analysis:}
This simple binary example already reveals a fundamental difference between predicting in causal and predicting in anticausal direction. The street will always be wet if there is rain $P(E = 1|C = 1) = 1$, no matter how likely it is to have rain. On the other hand, inferring there is rain based on the observation of a wet street highly depends on the likeliness of rain $P(C = 1|E = 1) = \frac{P(C = 1)}{N_E P(C = 0) + P(C = 1)}$. 

Comparing the conditionals \eqref{eq:causalMechanism} and \eqref{eq:anticausalMechanism} with respect to the risk of a misclassification, the error source in case of \eqref{eq:causalMechanism} is only the noise $N_E$, and in case of \eqref{eq:anticausalMechanism}, it is the product of $N_E$ and $P(C)$. Therefore, different risks of a misclassification can be expected depending on the causal prediction direction.

Motivated by this simple binary example, we want to identify the difference in the error source of predicting in causal and anticausal direction of regression problems. For this, we analyze the expected prediction error in terms of the expected loss when the true data generating function serves as prediction model. In the following, we consider $C$ and $E$ as continuous real valued variables and the squared error as loss function.

\subsection{Assumptions}
\setcounter{equation}{0}
Two further assumptions are particularly important for our analysis.

\subsubsection{Additive Noise Model}
We assume that data is generated by an \emph{additive noise model} (ANM) \cite{hoyer2009nonlinear}
\begin{equation}
E = \varphi(C, N_E) = \phi(C) + N_E,
\label{eq:ANM}
\end{equation}
where $C \in [0, 1]$, $N_E \in \mathbb{R}$ and $\phi: C \rightarrow \mathbb{R}$ is a deterministic \emph{oracle} function.\footnote{We call the function $\phi$ as ``oracle function'' in order to distinguish between the estimated function from a prediction model and the true function of the data generation process.} The cause is assumed to be normalized on $[0, 1]$ without loss of generality since this is only a matter of scaling. Cause and effect variables are assumed to share no unobserved common causes, which is also known as causal sufficiency assumption  \cite{Pearl:2009:CMR:1642718}. ANMs are particularly utilized for causal inference, where an ANM can only be fitted in the true causal direction aside from exceptions such as linear functions with Gaussian noise \cite{zhang2009identifiability}.

The inverse oracle function $\phi^{-1}$ is assumed to be defined on the whole domain of $E$. In practice, it is often assumed that $\operatorname{Var}[N_E]$ is much smaller compared to $\operatorname{Var}[\phi(C)]$.

An additive noise is generally a widely used assumption in prediction models that assume $Y = \widehat \phi(X) + N$, where the target $Y$ is influenced by additive noise $N$ and the goal is to learn the function $\widehat \phi$. Note that the differences between this additive noise assumption and ANM are the true causal relationships that are mostly carelessly ignored in prediction models. Nevertheless, the causal structure can have crucial implications for the prediction, and those are pointed out in this paper.

\subsubsection{Independence of mechanism and cause}
\label{sec:independence}
The \emph{independence of mechanism and cause} is the most crucial assumption concerning our work. Here, ``independence'' is different from the classical statistical definition, and it postulates that the cause distribution $p(C)$ is independent of the mechanism $p(E|C)$ and therefore has no information about it. In particular, changing $p(C)$ has no influence on $p(E|C)$ and vice versa. On the other hand, the effect distribution $p(E)$ may contain information about $p(E|C)$. This assumption is also related to the autonomous data generation process and exogeneity \cite{Pearl:2009:CMR:1642718,ZhaZhaSch15}.

Assuming the deterministic oracle function $\phi$ is a strictly monotonically increasing diffeomorphism\footnote{Diffeomorphism implies that the function is differentiable and has a differentiable inverse.}, the independence assumption can be formulated in terms of a positive dependency between the slope of $\phi$ and the distribution of the cause $p(C)$ \cite{JMLR:v16:janzing15a}.
\begin{post*}[Independence assumption]
If $C$ causes $E$ with $E = \phi(C) + N_E$ where $N_E {\perp\!\!\!\perp} C$ then
\begin{equation}
\operatorname{Dep}[\phi', p(C)] = 0,
\label{eq:indep}
\end{equation}
where $\phi'$ represents the derivative of $\phi$. This dependency can be analytically measured by a formula similar to the covariance, which characterizes the relation between two functions $\phi'(C)$ and $p(C)$
\begin{align}
\operatorname{Dep}[\phi', p(C)] & =  \int_0^1 \phi'(C) p(C) dC - \int_0^1 \phi'(C) dC \int_0^1 p(C) dC \nonumber \\
& = \int_0^1 \phi'(C) p(C) dC - (\phi(1) - \phi(0)) = 0.
\label{eq:cov1}
\end{align}
\end{post*}
Important to see here is that this formulation has no statistical properties and only represents a measure of the dependency between $\phi'$ and $p(C)$, where it might be helpful to rather see the density distribution $p(C)$ as a function of $C$ that is integrated over the support $[0, 1]$. Note that this postulate implies
\begin{equation*}
\int_0^1 \phi'(C) p(C) dC = \phi(1) - \phi(0).
\end{equation*}

Intuitively, if $\phi'$ is independent of $p(C)$, it is unlikely that a high slope of $\phi$ coincide with a high density of $p(C)$. \cite{JMLR:v16:janzing15a} showed that if this assumption holds and $\phi$ is not the identity function, then $\phi^{-1}{'}$ has a positive dependency with $p(E)$
\begin{equation*}
\operatorname{Dep}[\phi^{-1}{'}, p(E)] > 0,
\end{equation*}
which is illustrated in Figure \ref{fig:Slope}. A similar formulation is also possible by postulating that $p(C)$ and the $\operatorname{log}(\phi{'})$ are independent, which allows several information theoretic interpretations such as that $p(E)$ contains information about the mechanism $p(E|C)$ \cite{6620}.

The independence assumption seems plausible considering the aforementioned rain problem. The mechanism shown in Table \eqref{eq:causalMechanism} of ``generating'' a wet street is designed independently of it's input distribution $P(C)$. Therefore, changing the cause distribution $P(C)$, and thus the probability of observing rain, has no influence on this mechanism. As soon as it is raining, the street will be wet and the conditional will remain the same no matter how likely it is to observe rain at all. On the other hand, as the conditional in Table \eqref{eq:anticausalMechanism} indicates, observing a wet street highly depends on how likely rain is at all. If \eqref{eq:causalMechanism} changes, the effect distribution $P(E)$ changes but the cause distribution $P(C)$ remains the same. There is no dependency between the $P(E|C)$ and the cause distribution $P(C)$, but $P(C|E)$ depends on $P(C)$. These properties are well captured in the conditionals \eqref{eq:causalMechanism} and \eqref{eq:anticausalMechanism}. Therefore, the probabilities of a misclassification in causal and anticausal direction show an asymmetrical behavior.

\begin{figure}[t]
\centering
\includegraphics[width=0.5\columnwidth]{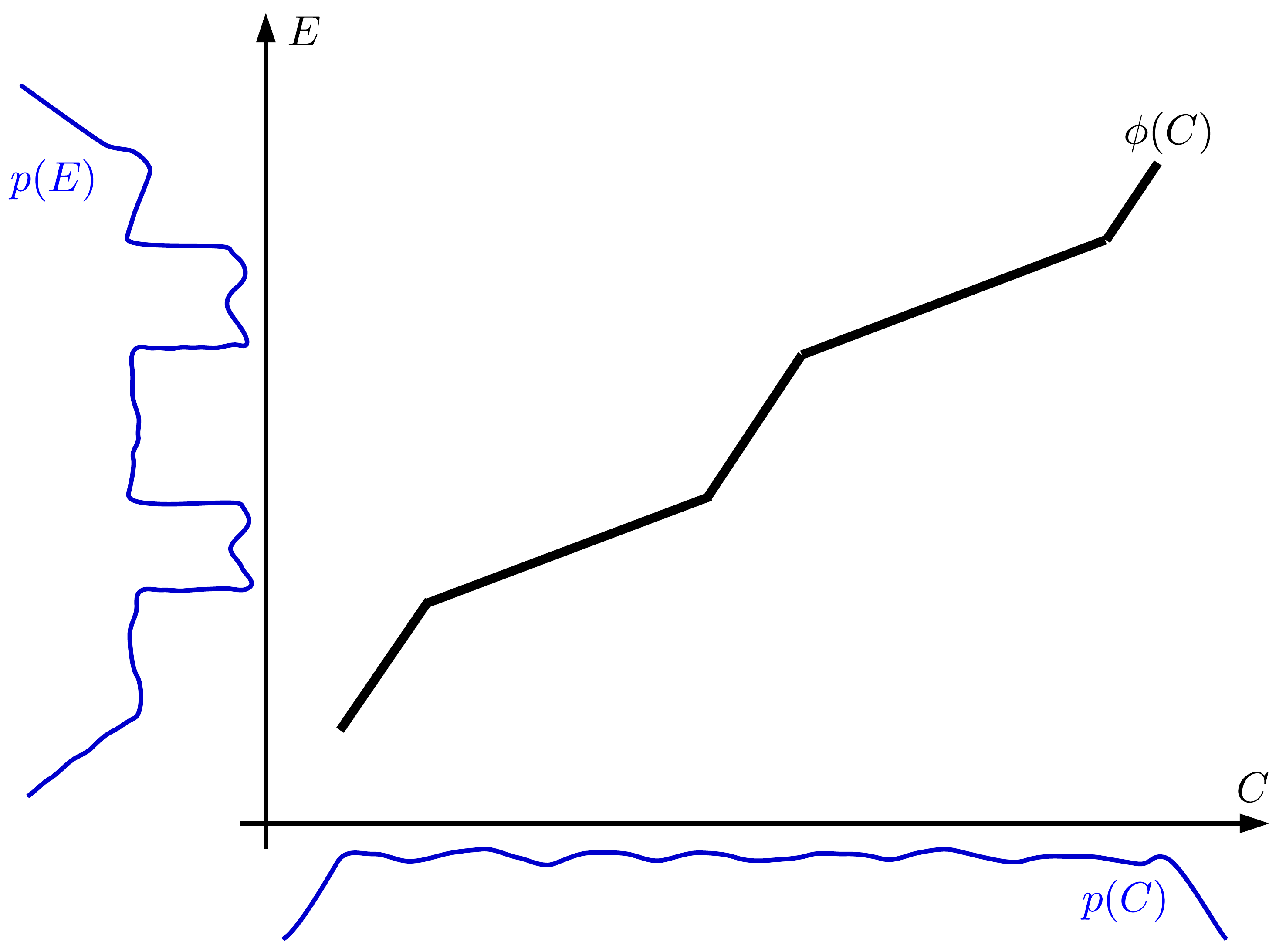}
\caption{Assuming $p(C)$ is independent of $\phi'$, then regions with high density in $p(E)$ coincide with regions where $\phi$ is flat. (Modified from \cite{janzing2012information})}
\label{fig:Slope}
\end{figure}

\section{Analysis of Regression Problems}
\label{ref:regprobs}
In the following, we analyze the implications of the ANM and independence assumption in causal and anticausal regression problems in terms of the expected prediction error. In an anticausal prediction problem, the goal is to learn the inverse oracle function $\phi^{-1}$ in order to predict the cause $C$ from the effect $E$ since we assume that $\phi$ represents the true relationship between these variables. According to \eqref{eq:ANM}, $E$ is influenced by noise, and hence, the predicted cause will be noisy too. This already leads to three important aspects. First, the prediction of $C$ is also affected by the noise of the effect and thus ordinary regression techniques, such as least squares regression, will fail to accurately predict the noise free cause that generated the data. Second, $\phi$ needs to be invertible for accurate anticausal predictions. If $\phi$ is not injective for restricted domains or surjective for any domains, there is a general information loss by predicting in anticausal direction. Third, a more accurate estimation of $\phi^{-1}$ in anticausal direction may be obtained by inverting $\phi$ in causal direction. The latter is further discussed in Section \ref{sec:optimalDesignInverse}. As already mentioned in Section \ref{sec:independence}, we assume that $\phi$ is a strictly monotonically increasing diffeomorphism which is convenient for the analysis, but according to \cite{JMLR:v16:janzing15a}, the diffeomorphism assumption can be significantly weakened by constraining $\phi$ to be almost everywhere differentiable such as in Figure \ref{fig:Slope}.

Regarding the intrinsic structure of a prediction problem, an optimal prediction is only possible if the oracle function $\phi$, which captures this structure, is used as estimator. Due to the natural variability of a system, such as stochastic noise influencing the effect, it is rarely possible to infer the oracle function. Further, even if the oracle function is used, the noise represents an irreducible error.

In the following, we focus the discussion on the expected prediction error when the oracle function and its inverse are utilized for the prediction.

\subsection{Expected Prediction Error}
\label{sec:expectedPredError}
Given an estimator $\widehat \phi$, an input value $x$ and a target value $y$, the prediction quality of the estimator can be measured by a loss function $L(y, \widehat \phi(x)) \geq 0$, where the squared error $L_\text{SE}(y, \widehat \phi(x)) = (y - \widehat \phi(x))^2$ is a typical loss function for regression problems. The expected prediction error is given by the expectation of the loss function
\begin{align*}
\mathbb{E}[L(Y, \widehat \phi(X))] = & \int \int L(Y, \widehat \phi(X)) p(X) p(Y|X) dX \ dY \nonumber \\
= & \int \int L(Y, \widehat \phi(X)) p(X, Y) dX \ dY,
\end{align*}
where it is implicitly assumed that the joint distribution $p(X, Y)$ reflects the underlying problem. The densities $p(X)$ and $p(Y|X)$ can be seen as weighting factors of the error. Intuitively, the error is higher weighted if it is more likely to observe $x$ (regions with high density in $p(X)$) and, further, if it is more likely to observe $y$ given $x$ (regions with high density in $p(Y|X)$).

\subsection{Optimal design for inverse predictions: Inverse vs Reverse regression}
\label{sec:optimalDesignInverse}
\begin{figure}[t]
\subfigure[$\phi(X) = X$]{
  \centering
  \includegraphics[width=0.31\columnwidth]{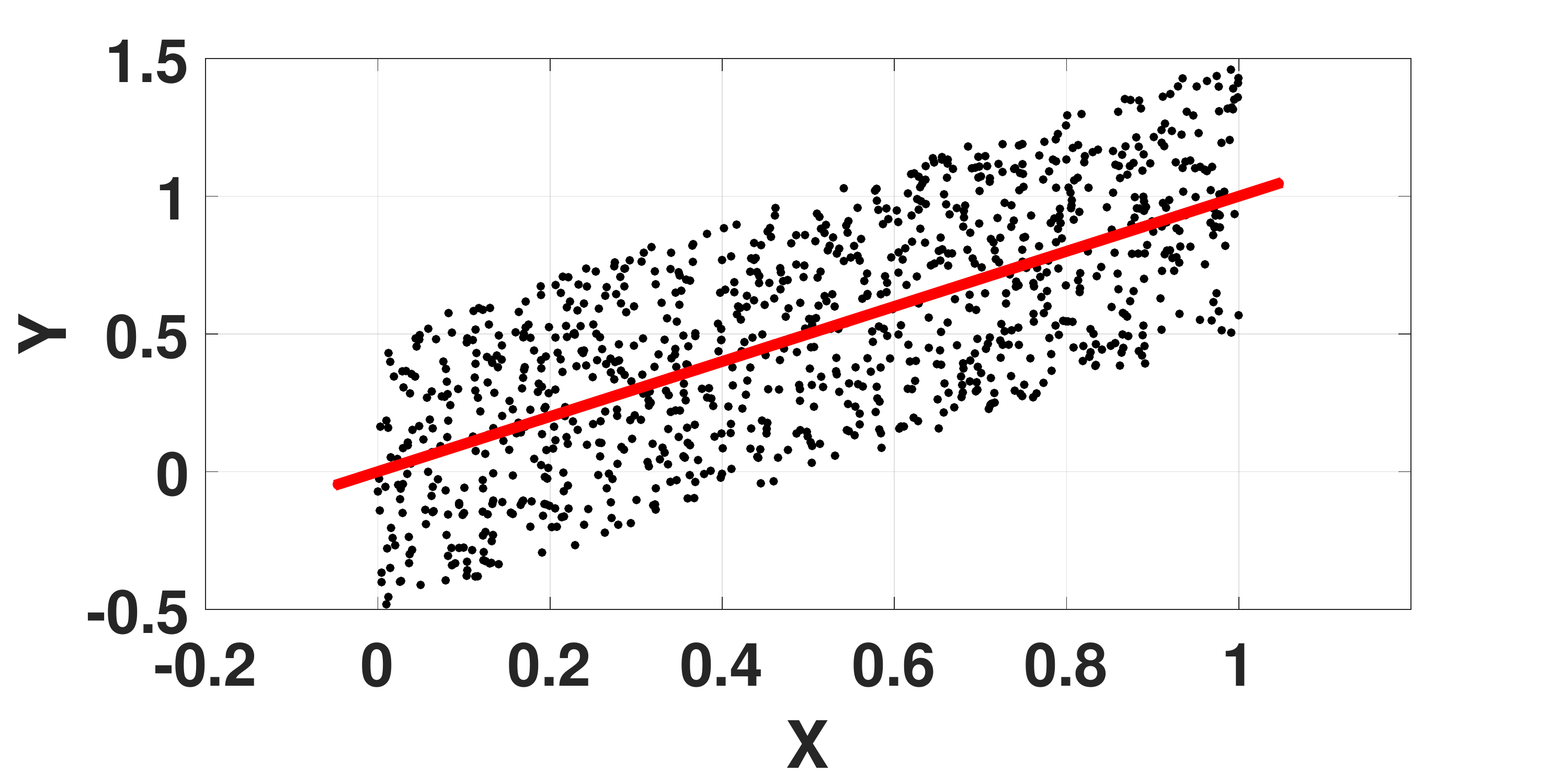}
}
\subfigure[$\widehat{\phi^{-1}}(Y) = \frac{Y}{2} + 0.25$]{
  \centering
  \includegraphics[width=0.31\columnwidth]{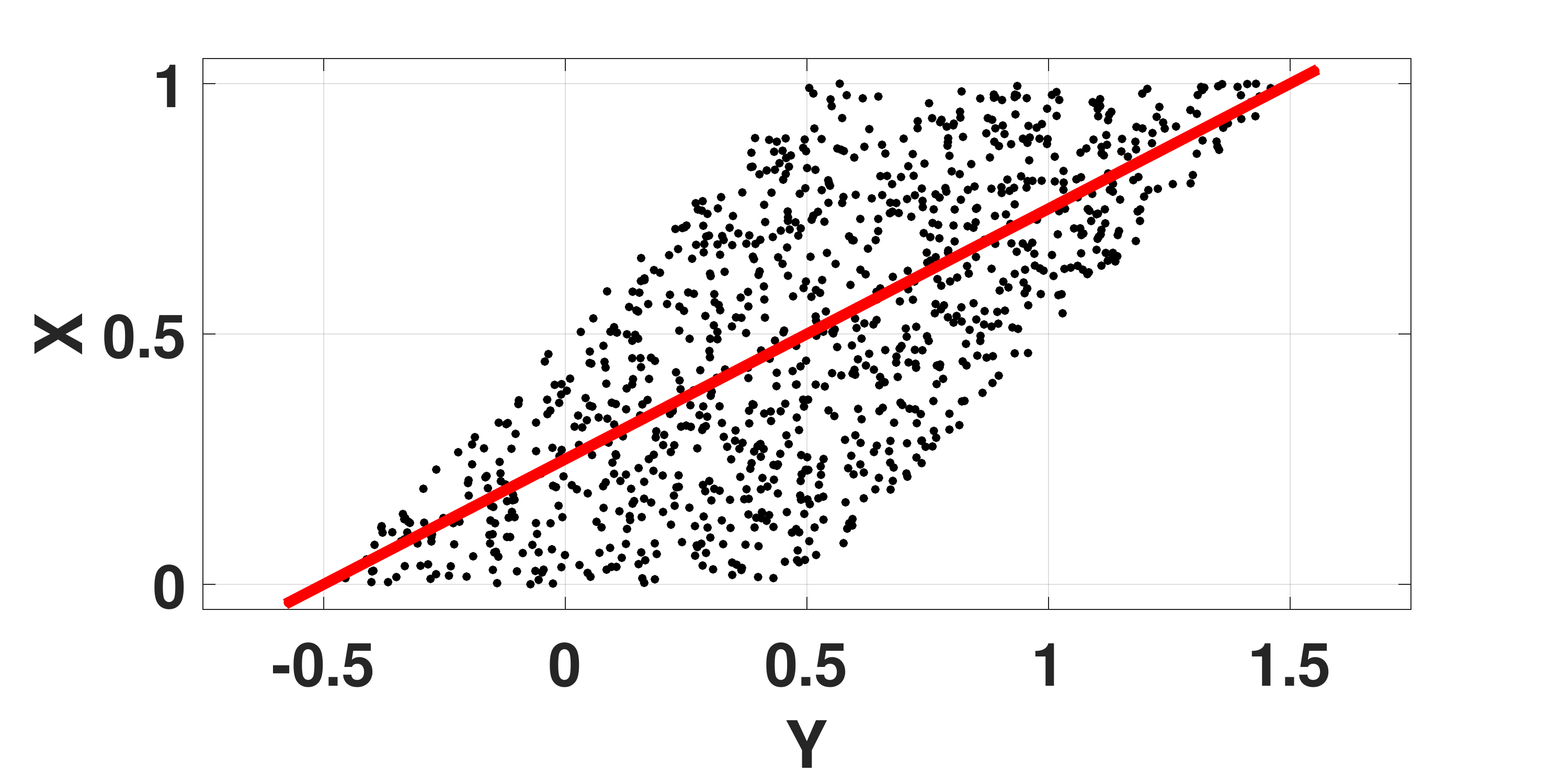}
}
\subfigure[$\phi^{-1}(Y) = Y$]{
  \centering
  \includegraphics[width=0.31\columnwidth]{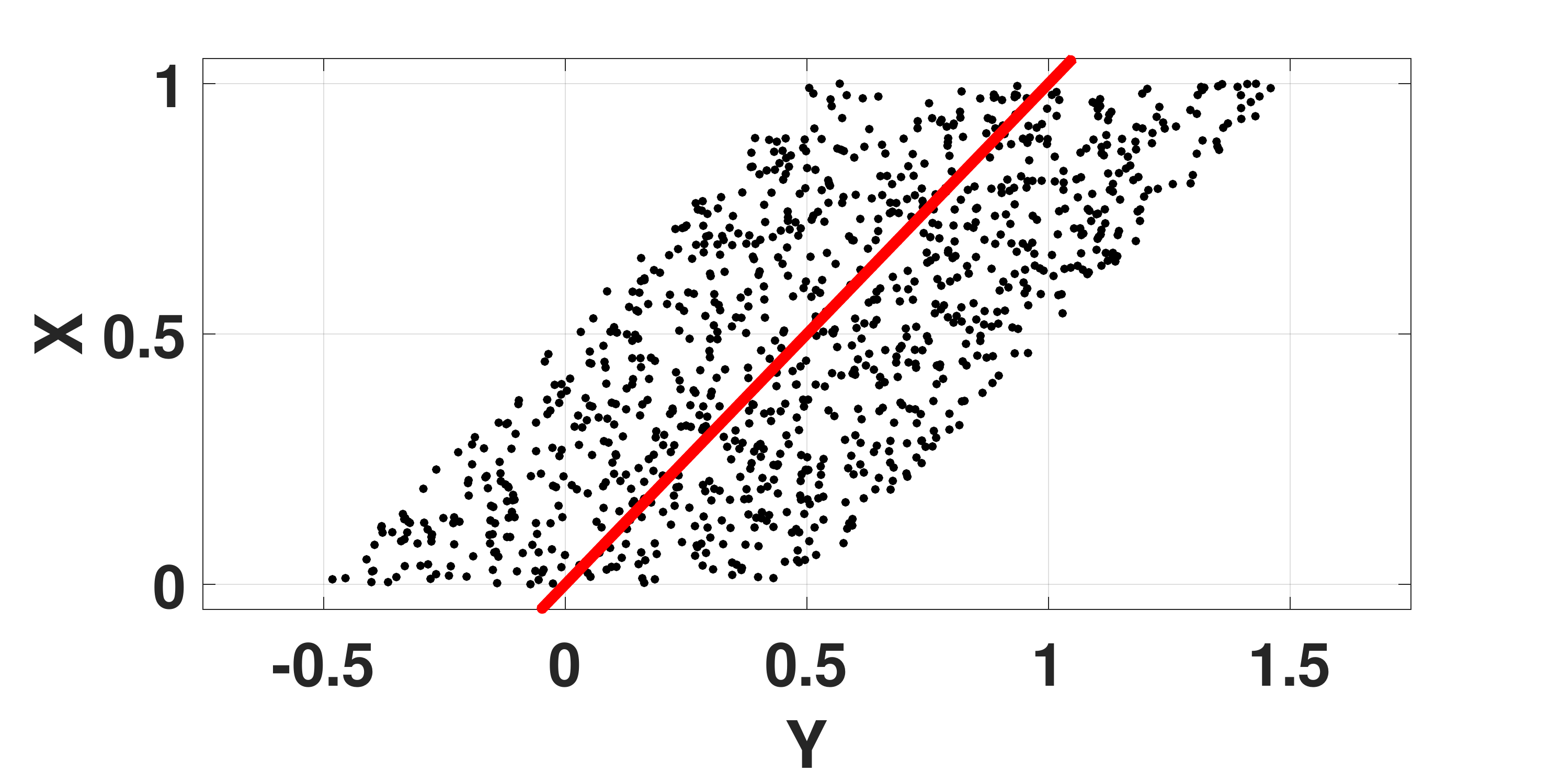}
}
\caption{A simple example where the least squares solution for an inverse prediction is highly biased and does not represent $\phi^{-1}$. \textbf{(a)} A regression problem defined by $Y = \phi(X) + N_Y = X + N_Y$, where $X \sim U(0, 1)$ and $N_Y \sim U(-0.5, 0.5)$. The true prediction model is given by $\phi(X) = Y$, which also coincides with the least squares solution. \textbf{(b)} The corresponding least squares solution $\widehat{\phi^{-1}}$ for the inverse prediction problem. This prediction model does not reflect the inverse relationship given by the true $\phi^{-1}$. \textbf{(c)} The true prediction model for an inverse prediction of $Y$ given by $\phi^{-1}$.}
\label{fig:optimalDesign}
\end{figure}
In many applications, the prediction of the cause rather than the effect is of interest. For instance, predicting the disease that caused a certain symptom. In these anticausal prediction scenarios, so called \emph{calibration models} provide several techniques to predict the cause based on the effect \cite{lavagnini2007statistical}. The idea of many calibration models is to fit a model from $C$ to $E$ and invert it in order to obtain a model for the anticausal direction. This is referred as "\emph{inverse regression}". On the other hand, one could also simply fit a regression model from $E$ to $C$, which is referred as "\emph{reverse regression}" or sometimes "direct regression" \cite{parker2010prediction}.

However, there is a big difference between inverse and reverse regression. The literature regarding this comparison is somewhat underdeveloped, but an understanding of the difference is important. The work of \cite{parker2010prediction,kuo2012comparison} provide a comparison between inverse and reverse regression, where they point out that the solution of reverse regression is more biased than the solution of inverse regression. While many data scientists naively perform a reverse regression due to the simplicity or due to the lack of awareness of the difference, an inverse regression should be preferred in anticausal prediction problems to minimize the risk of wrong conclusions.

Seeing this, we use an inverse regression for anticausal predictions and provide thereby a new contribution to the comparison. In general, the relation between cause and effect is according to the ANM and independence assumption defined by the oracle function $\phi$, and thus, $\phi^{-1}$ reflects the inverse relation. This is obvious in the deterministic case without noise
\begin{equation*}
\phi^{-1}(Y) = \phi^{-1}(\phi(X)) = X.
\end{equation*}
However, in the ANM, we assume additive noise on the effect $E = \phi(C) + N_E$. In this case, inferring the causal oracle function $\phi$ defines the general problem in regression tasks and can, for instance, be solved by minimizing a loss function. For example, the least squares solution in causal direction coincides with $\phi$
\begin{equation*}
\underset{\widehat \phi}{\operatorname{arg\,min\, }}\mathbb{E}[(E - \widehat \phi(C))^2] = \underset{\widehat \phi}{\operatorname{arg\,min\, }}\mathbb{E}[(\phi(C) + N_E - \widehat \phi(C))^2] = \phi.
\end{equation*}
On the contrary, in case of a reverse regression in anticausal direction, the inverse oracle function $\phi^{-1}$ may not coincide with the least squares solution due to the noise
\begin{equation}
\underset{\widehat{\phi^{-1}}}{\operatorname{arg\,min\, }}\mathbb{E}[(C - \widehat{\phi^{-1}}(\phi(C) + N_E))^2] \neq \underset{\widehat{\phi^{-1}}}{\operatorname{arg\,min\, }}\mathbb{E}[(C - \widehat{\phi^{-1}}(\phi(C)))^2].
\label{eq:notEqual}
\end{equation}
A simple example would be $E = \phi(C) + N_E = C + N_E$, where $C \sim U(0, 1)$ and $N_E \sim U(-0.5, 0.5)$ as illustrated in Figure \ref{fig:optimalDesign}. The least squares solution for the reverse estimator is given by $\widehat{\phi^{-1}}(E) = \frac{E}{2} + 0.25$, which is clearly different from the true inverse relationship defined by $\phi^{-1}(E) = E$. For instance, the desired optimal inverse prediction for $E = 1$ would be $\phi^{-1}(1) = 1$, but the least squares solution gives $\widehat{\phi^{-1}}(1) = 0.75$, which is highly biased. Note that the smaller the variance of the noise, the closer the least squares solution is to the inverse function
\begin{equation}
\underset{\operatorname{Var}[N_E] \rightarrow 0}{\operatorname{lim}}\underset{\widehat{\phi^{-1}}}{\operatorname{arg\,min\, }}\mathbb{E}[(C - \widehat{\phi^{-1}}(\phi(C) + N_E))^2] = \phi^{-1}.
\label{eq:limInverse}
\end{equation}
Seeing this, an inverse regression by first estimating the causal function $\phi$ and then invert it in order to obtain an estimation of the inverse function $\phi^{-1}$ as suggested in \cite{franccois2004optimal,lavagnini2007statistical} is clearly the better choice. This also shows that knowing the causal direction can provide useful information for the prediction and may reduce the risk of a fatal decision for a patient treatment due to a misinterpretation of the prediction results.

As mentioned before, we will, thus, consider inverse regression instead of reverse regression for anticausal predictions and show in the next section that, by using the true function $\phi$, the expected error of a causal prediction is hence different from the expected error of an anticausal prediction due to the additive noise. This difference is captured by the expected least squared error of causal and anticausal predictions.

\subsection{Expected prediction error of the oracle function}
In the following, the squared error is used as loss function. The expected error in causal problems is given by
\begin{align*}
	\mathcal{E}_{E|C} = & \int \int_0^1 (E - \widehat \phi(C))^2  p(C, E) dC \ dE \nonumber \\
	= & \int \int_0^1 (\phi(C) + N_E - \widehat \phi(C))^2 p(C) p(N_E) dC \ dN_E,
\end{align*}
where the subscript $\cdot_{E|C}$ indicates that the effect is predicted from the cause. Here, we used the additive noise assumption in terms that $C$ is independent of $N_E$ and that the joint density of $p(C, E)$ is equal to $p(C, N_E)$ since $\phi(C)$ is deterministic
\begin{equation*}
p(C, E) = p(C, \phi(C) + N_E) = p(C, N_E) = p(C) p(N_E).
\end{equation*} 
Following this, the expected error of the oracle function, which is the lower bound on the causal prediction, is given if $\widehat \phi = \phi$
\begin{align}
	\mathcal{E}_{E|C} = & \int \int_0^1 (\phi(C) + N_E - \phi(C))^2 p(C) p(N_E) dC \ dN_E \nonumber \\
	=& \int \int_0^1 N_E^2 p(N_E) p(C) dC \ dN_E \nonumber \\
	=& \operatorname{Var}[N_E].
	\label{eq:ExCas}
\end{align}
The lower bound on the expected error therefore only depends on the variance of the error noise.

In case of the anticausal prediction via inverse regression, the expected error is given by
\begin{align*}
	\mathcal{E}_{C|E} = & \int \int_0^1 (C - \widehat{\phi^{-1}}(E))^2  p(C, E)dC \ dE \nonumber \\
	= & \int \int_0^1 (C - \widehat{\phi^{-1}}(\phi(C) + N_E))^2 p(C) p(N_E) dC \ dN_E,
\end{align*}
where $\widehat{\phi^{-1}}$ represents the inverse prediction model. As argued in Section \ref{sec:optimalDesignInverse}, the expected error of the inverse oracle function in anticausal problems is similarly given if $\widehat{\phi^{-1}} = \phi^{-1}$
\begin{align}
	\mathcal{E}_{C|E} = & \int \int_0^1 (C - \phi^{-1}(\phi(C) + N_E))^2 p(C) p(N_E) dC \ dN_E \nonumber \\
	\approx & \int \int_0^1 \left(C - \phi^{-1}(\phi(C)) + N_E \phi^{-1}{'}(\phi(C))\right)^2 p(C) p(N_E) dC \ dN_E \nonumber \\
	= & \int \int_0^1 \left(C - \phi^{-1}(\phi(C)) + N_E \frac{1}{\phi'(\phi^{-1}(\phi(C)))}\right)^2 p(C) p(N_E) dC \ dN_E \nonumber \\
	= & \int \int_0^1 N_E^2 p(N_E) \left(\frac{1}{\phi'(C)}\right)^2 p(C) dC \ dN_E \nonumber \\
	= & \operatorname{Var}[N_E] \int_0^1 \left(\frac{1}{\phi'(C)}\right)^2 p(C) dC,
	\label{eq:ExAnti}
\end{align}
where we, similar as in the work by \cite{bishop1995training}, approximate $\phi^{-1}(\phi(C) + N_E)$ by a Taylor expansion
\begin{equation*}
\phi^{-1}(\phi(C) + N_E) = \phi^{-1}(\phi(C)) + N_E \frac{1}{\phi{'}(C)} + \mathcal{O}(N_E^2),
\end{equation*}
under the assumption of sufficiently small noise such that the rest error $\mathcal{O}(N_E^2)$ can be neglected, which is a natural assumption in regression problems. We further use that $f^{-1}{'}(X) = \frac{1}{f{'}(f^{-1}(X))}$ since $\phi$ is a strictly monotonically increasing diffeomorphism. Note that \eqref{eq:ExAnti} represents the lower bound on the expected prediction error for inverse regression, but does not necessarily coincide with the lower bound of reverse regression.

Comparing \eqref{eq:ExCas} and \eqref{eq:ExAnti} already reveals a fundamental difference; the expected error in causal prediction is independent of $\phi$, but in anticausal predictions, it heavily depends on $\phi{'}$. However, a further statement about $\int_0^1 \left(\frac{1}{\phi'(C)}\right)^2 p(C) dC$ is not possible at this point.

So far, we did not use the independence assumption \eqref{eq:indep}, but it can now be used to say something about $\int_{0}^{1} \left(\frac{1}{\phi'(C)}\right)^2 p(C) dC$. By applying the Cauchy-–Schwarz inequality we first conclude that
\begin{align*}
 \int_0^1 \left(\frac{1}{\phi'(C)}\right)^2 p(C) dC = & \int_0^1 \left(\frac{1}{\phi'(C)}\right)^2 p(C) dC \underbrace{\int_0^1 1^2 p(C) dC}_{= \ 1} \nonumber \\
\geq  & \left(\int_0^1 \frac{1}{\phi'(C)} p(C) dC \right)^2.
\end{align*}
Note that the densities $p(C)$ are strictly positive. The independence assumption formalized in \eqref{eq:cov1} states $\int_0^1 \phi'(C) p(C) dC = \phi(1) - \phi(0)$. Seeing this, it can be concluded that
\begin{align*}
\int_0^1 \frac{1}{\phi'(C)} p(C) dC = & \int_0^1 \frac{1}{\phi'(C)} p(C) dC \cdot\underbrace{\frac{\int_0^1 \phi'(C) p(C) dC}{\phi(1) - \phi(0)}}_{=\ 1} \nonumber \\
= & \frac{\int_0^1 \left(\sqrt{\frac{1}{\phi'(C)}}\right)^2 p(C) dC \int_0^1 \left(\sqrt{\phi'(C)}\right)^2 p(C) dC}{\phi(1) - \phi(0)} \nonumber \\
\geq & \frac{\left(\int_0^1 \frac{\sqrt{\phi'(C)}}{\sqrt{\phi'(C)}} p(C) dC\right)^2}{\phi(1) - \phi(0)} = \frac{1}{\phi(1) - \phi(0)}.
\end{align*}
Combining this result with the previous one gives
\begin{align*}
\int_0^1 \left(\frac{1}{\phi'(C)}\right)^2 p(C) dC \geq & \left(\underbrace{\int_0^1 \frac{1}{\phi'(C)} p(C) dC}_{\geq \ \frac{1}{\phi(1) - \phi(0)}}\right)^2 \nonumber \\
\geq & \ \frac{1}{(\phi(1) - \phi(0))^2}.
\end{align*}
The final conclusion of \eqref{eq:ExAnti} is therefore
\begin{equation}
	\mathcal{E}_{C|E} \approx \operatorname{Var}[N_E] \underbrace{\int_0^1 \left(\frac{1}{\phi'(C)}\right)^2 p(C) dC}_{\geq \ \frac{1}{(\phi(1) - \phi(0))^2}}.
\label{eq:antiFinal}
\end{equation}
Now, the relationship between \eqref{eq:ExCas} and \eqref{eq:antiFinal} depends on the scaling of $\phi$, where it can be distinguished between two cases:
\begin{enumerate}
	\item If $\phi(1) - \phi(0) \leq 1$ then $\int_0^1 \left(\frac{1}{\phi'(C)}\right)^2 p(C) dC \geq 1$\\
	\item If $\phi(1) - \phi(0) \geq 1$ then $\int_0^1 \left(\frac{1}{\phi'(C)}\right)^2 p(C) dC > 0$
\end{enumerate}
Note that $\int_0^1 \left(\frac{1}{\phi'(C)}\right)^2 p(C) dC = 1$ is only given in the case where $\phi(1) - \phi(0) = 1$ and $\phi$ is linear. Also, \eqref{eq:antiFinal} becomes strictly equal in the linear case. If $\phi$ is non-linear and $\phi(1) - \phi(0) \leq 1$ the expression becomes greater than $1$. An alternative interpretation of \eqref{eq:antiFinal} in the first case is that $\phi$ minimizes the squared error in causal direction with respect to the noise, but $\phi^{-1}$ does not guarantee to minimize the expected squared error in the anticausal direction as stated in \eqref{eq:notEqual}. Therefore, the additive noise has a higher negative impact on the expected error of anticausal predictions as compared to causal predictions. Note that this is not due to the choice of the squared error as loss function, but to the intrinsic relation between $\phi$ and $\phi^{-1}$, which is captured by the squared error.

This result leads to the following fundamental theorem:
\begin{theorem}[Error asymmetry]
\label{eq:theorem}
Let $C \in [0, 1]$ and $\phi: C \rightarrow \mathbb{R}$ be a strictly monotonically increasing diffeomorphism with $\phi(1) - \phi(0) \leq 1$. If \eqref{eq:ANM} and \eqref{eq:indep} hold, then the expected error of the oracle function $\phi$ in causal regression is smaller than or equal to the expected error of the inverse oracle function $\phi^{-1}$ in anticausal regression.
\begin{equation}
\phi(1) - \phi(0) \leq 1 \Rightarrow \mathcal{E}_{E|C} \leq \mathcal{E}_{C|E}
\end{equation}
where $\mathcal{E}_{E|C} = \mathcal{E}_{C|E}$ only if $\phi(1) - \phi(0) = 1$ and $\phi$ is linear. Further, $\mathcal{E}_{E|C} < \mathcal{E}_{C|E}$ if $\phi(1) - \phi(0) < 1$.
\end{theorem}
\textbf{Proof} \ This directly follows by comparing \eqref{eq:ExCas} and \eqref{eq:antiFinal}. \begin{flushright}
$\square$
\end{flushright}
If the assumptions hold, the theorem states an asymmetry of the prediction error with respect to whether the cause is predicted from the effect or the effect from the cause w.r.t. to the data generating function. As an extension to the implications of the independence postulate, the theorem implies a positive dependency between the slope of $\phi$ and the error in anticausal prediction, but not in causal prediction. This asymmetry has theoretical and practical implications in various domains. Some are briefly discussed in Section \ref{sec:contributions}.

This theorem has an important implication:
\begin{corollary}[Error asymmetry in normalized data]
\label{eq:normalizedData}
Let $C$ and $E$ be normalized to $[0, 1]$ and $\operatorname{Var}[N_E] > 0$. If \eqref{eq:ANM} and \eqref{eq:indep} hold, then the expected error of the oracle function $\phi$ in causal regression is strictly smaller than the expected error of the inverse oracle function $\phi^{-1}$ in anticausal regression.
\begin{equation*}
\mathcal{E}_{E|C} < \mathcal{E}_{C|E}
\end{equation*}
\end{corollary}
According to the additive noise assumption, the normalized effect is defined by 
\begin{align*}
	E & := \frac{E - E_\text{min}}{E_\text{max} - E_\text{min}} \\
	& = \frac{\phi(C) + N_E - (\phi(C) + N_E)_\text{min}}{(\phi(C) + N_E)_\text{max} - (\phi(C) + N_E)_\text{min}},
\end{align*}
where $E_\text{min} := (\phi(C) + N_E)_\text{min}$ and $E_\text{max} := (\phi(C) + N_E)_\text{max}$ denote the domain minimum and domain maximum of $E$, respectively. In practice, this is approximately given by normalizing the observed effect data. Due to the noise, the range of $\phi$ becomes $\phi(1) - \phi(0) < 1$. Note that this corollary is particularly interesting seeing that a smaller error in causal regression can be expected if the cause and effect data have the same scaling.

In the next section, we evaluate the theorem in artificial and real-world data sets where the cause and effect are known in advance.

\section{Empirical Evaluations}
\label{sec:experiments}
We first evaluated Theorem \ref{eq:theorem} with artificial data sets that fulfill the independence and additive noise assumption. For this, we performed evaluations in two different settings. In the first setting, it is assumed that the oracle function $\phi$ and its inverse are known. In the second setting, the causal direction and the oracle function $\phi$ are assumed to be unknown. Further, we used real-world data sets to see how robust the made assumptions are and how general the theorem is. In general, the root mean squared errors (RMSE) of predicting in causal and anticausal direction were compared. If the theorem holds, the RMSE of predicting in causal direction should be smaller or equal to the RMSE of predicting in anticausal prediction.

\subsection{Artificial data sets}
For all artificial data sets, we generated uniformly distributed cause data with values in $[0, 1]$ and chose functions $f$ that are monotonically increasing diffeomorphisms. The oracle function $\phi$ is normalized in terms of attaining values in $[0, 1]$ and the parameters of $\phi^{-1}$ are defined by $\phi$. The additive noise is Gaussian distributed. Therefore, cause data $C$, noise data $N_E$ and effect data $E$ were generated in the following way:
\begin{align*}
& C \sim \ U(0, 1) \text{ with } \operatorname{min}[C] = 0, \operatorname{max}[C] = 1 \\
& N_E \sim N(0, \sigma) \\
& \phi(C) = \frac{f(C) - f(0)}{f(1) - f(0)} \\
& E = \phi(C) + N_E
\end{align*}
As function $f$ we used $f(x) = \operatorname{exp}(a \cdot x)$, $f(x) = \operatorname{sin}\left(\frac{(30 \cdot x + 25) \cdot 2\pi}{360}\right)$ and $f(x) = x^a$. In all cases, we took care that the additive noise does not change the values of $E$ to invalid values with respect to the inverse $\phi^{-1}$.

\subsubsection{$\phi$ and the causal direction are known}
\begin{table}[t]
\caption{Comparison of the RMSE of predicting in causal and anticausal direction with linear $\phi$.}
\label{tabel:linear}
\center
\begin{tabular}{|c|c|c|c|}
\hline 
 & $\sigma = 0$ & $\sigma = 0.1$ & $\sigma = 0.5$ \\ 
\hline 
$\text{RMSE}_{E|C}$ & $0$ & $0.1 \pm 0.0002$ & $0.5 \pm 0.0011$ \\ 
\hline 
$\text{RMSE}_{C|E}$ & $0$ & $0.1 \pm 0.0002$ & $0.5 \pm 0.0011$ \\ 
\hline 
\end{tabular} 
\end{table}

\begin{figure}[t]
\subfigure[$\operatorname{exp}(x)$]{
  \centering
  \includegraphics[width=0.23\columnwidth]{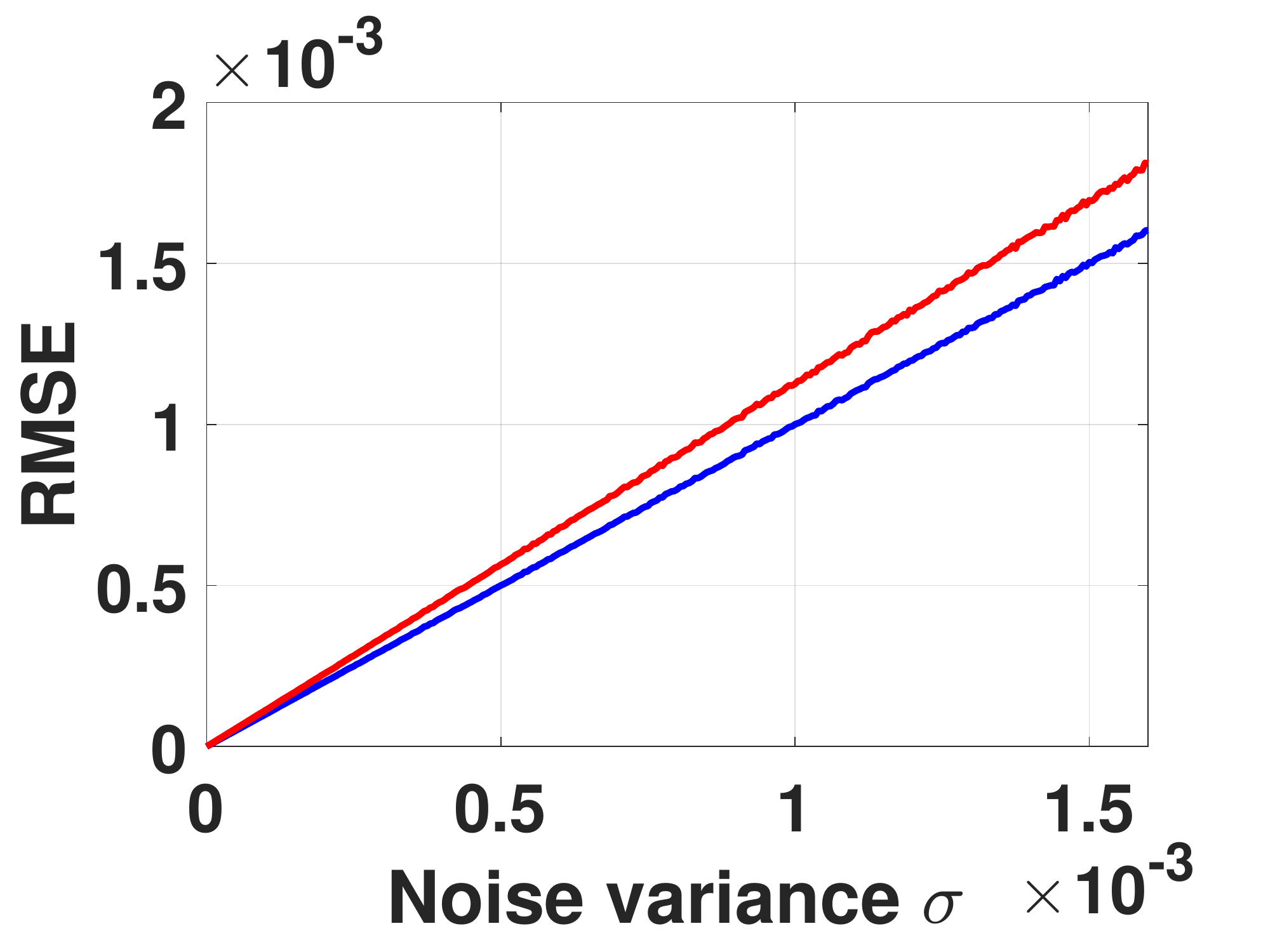}
  \label{fig:exp11}
}
\subfigure[$\operatorname{exp}(2 \cdot x)$]{
  \centering
  \includegraphics[width=0.23\columnwidth]{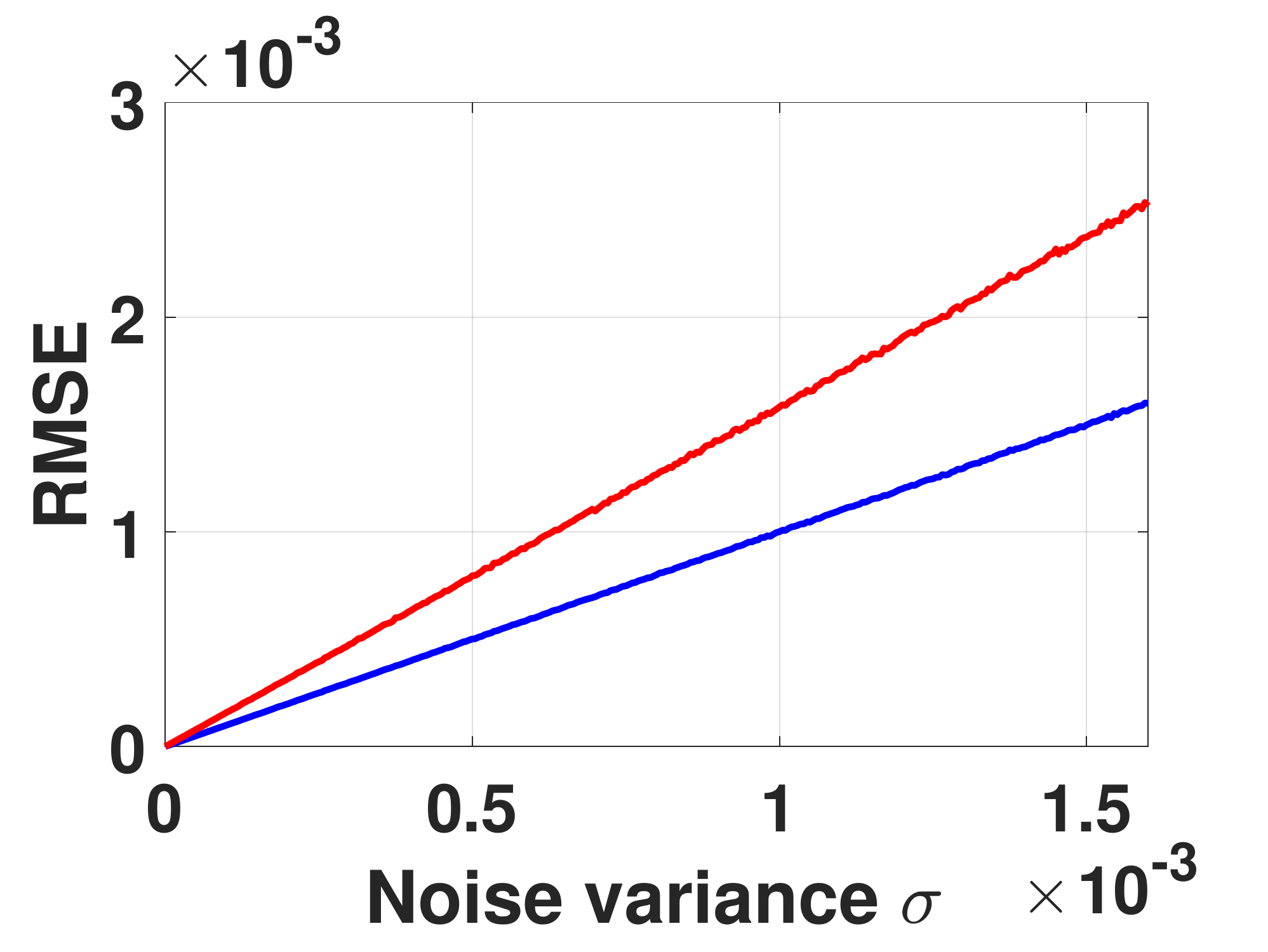}
  \label{fig:exp12}
}
\subfigure[$\operatorname{exp}(5 \cdot x)$]{
  \centering
  \includegraphics[width=0.23\columnwidth]{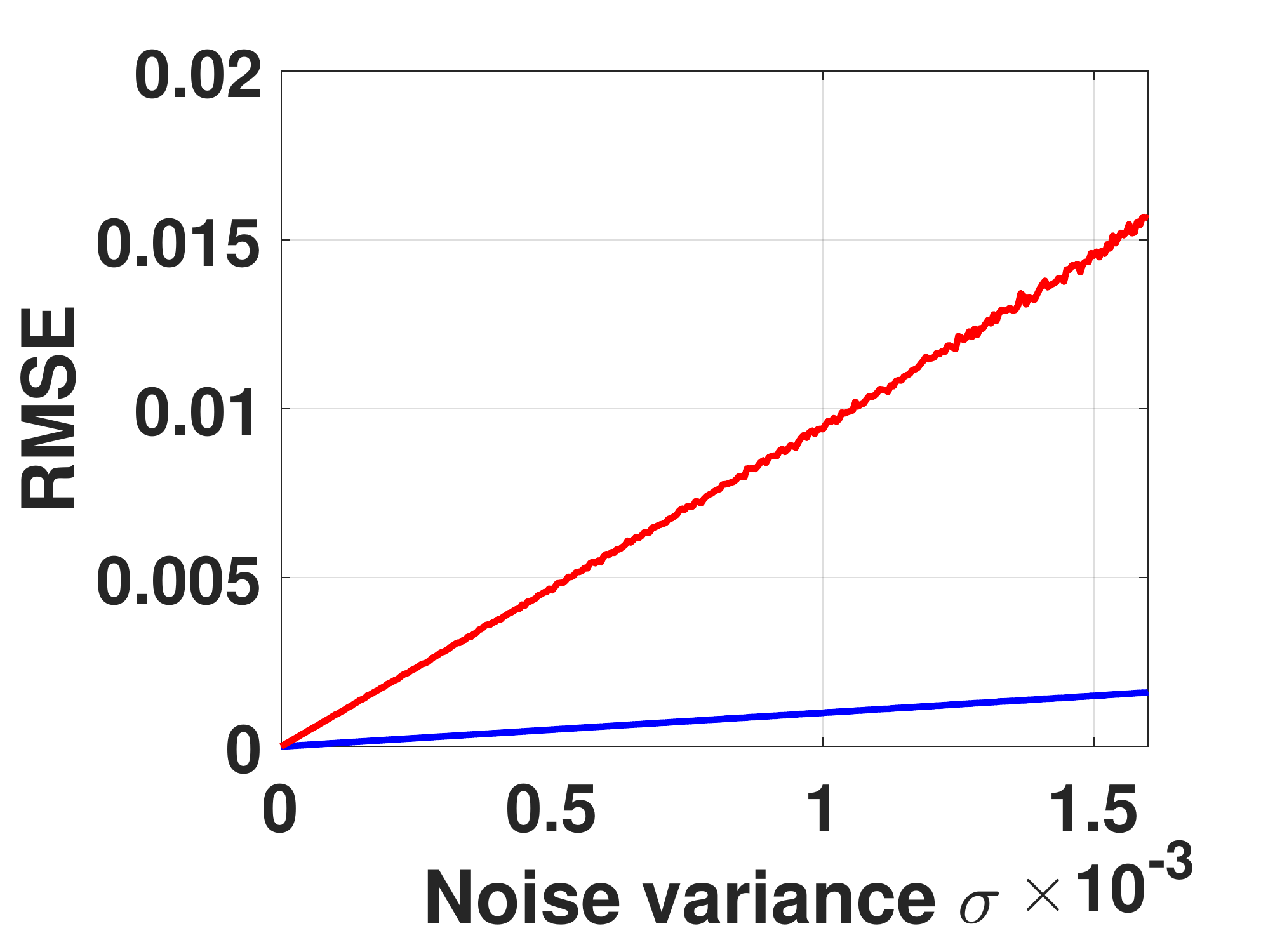}
  \label{fig:exp15}
}
\subfigure[$\operatorname{sin}\left(\frac{(30 \cdot x + 25) \cdot 2\pi}{360}\right)$]{
  \centering
  \includegraphics[width=0.23\columnwidth]{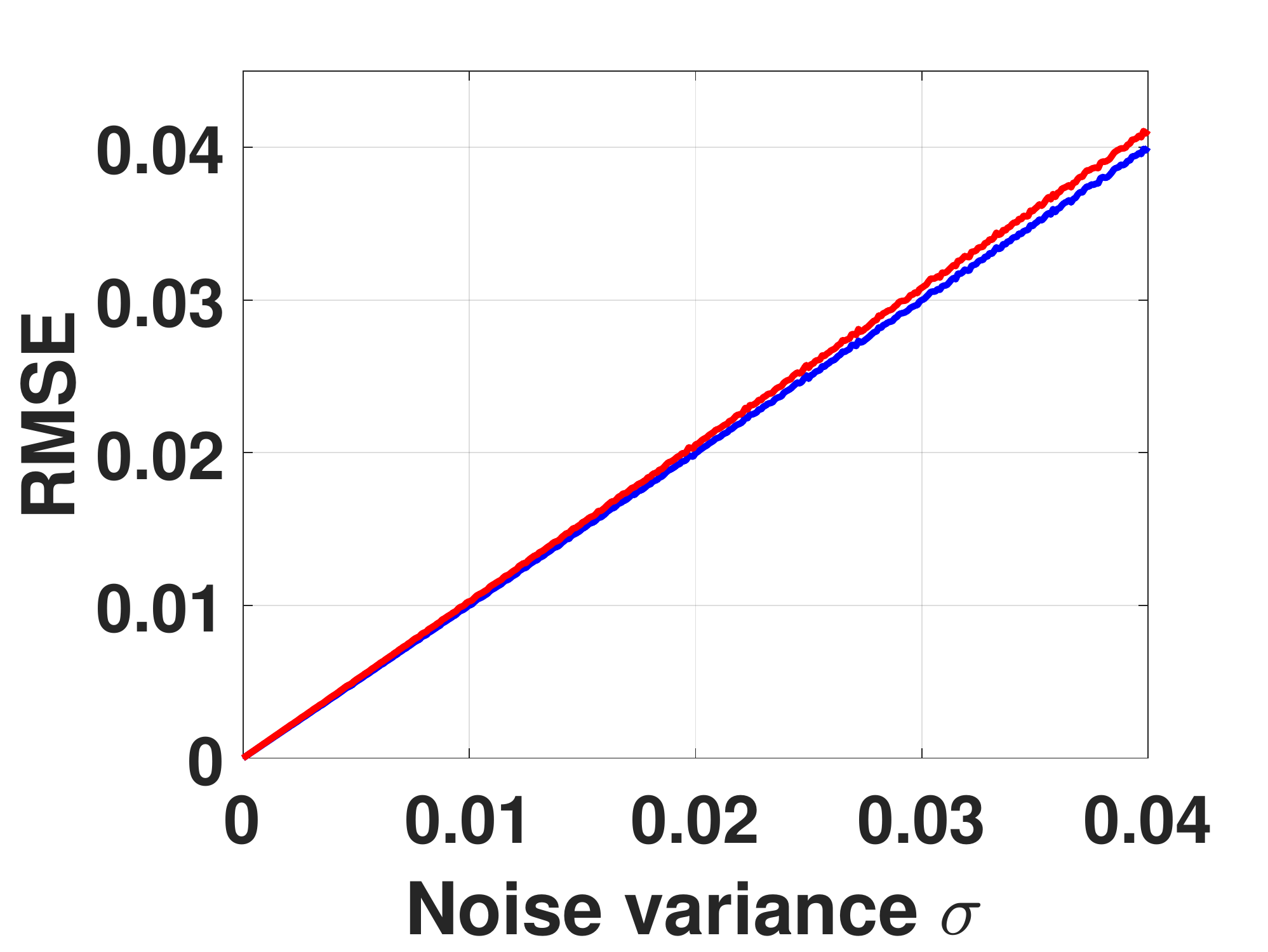}
  \label{fig:sinnd}
}
\subfigure[$x^{0.2}$]{
  \centering
  \includegraphics[width=0.23\columnwidth]{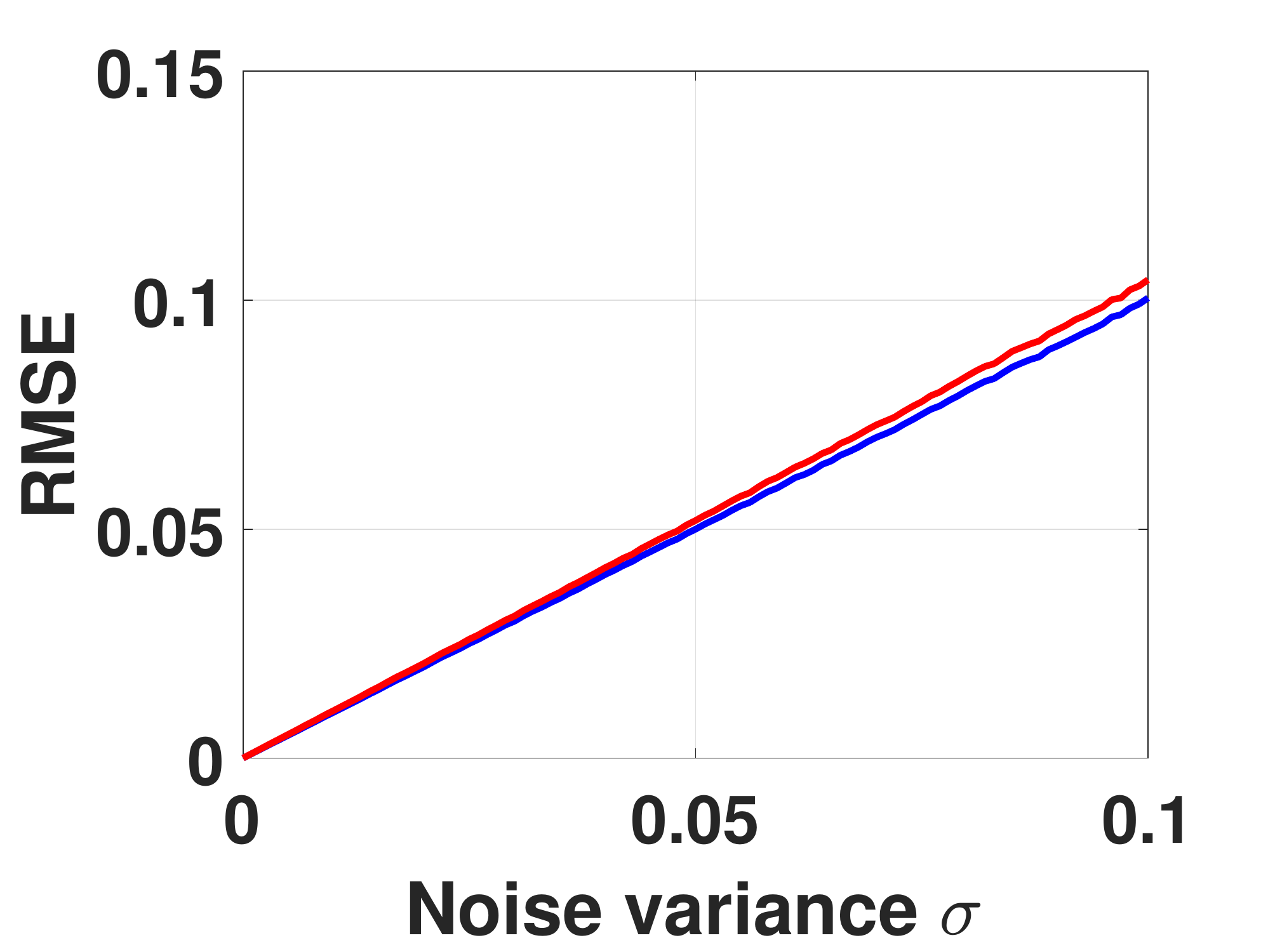}
  \label{fig:x02}
}
\subfigure[$x^2$]{
  \centering
  \includegraphics[width=0.23\columnwidth]{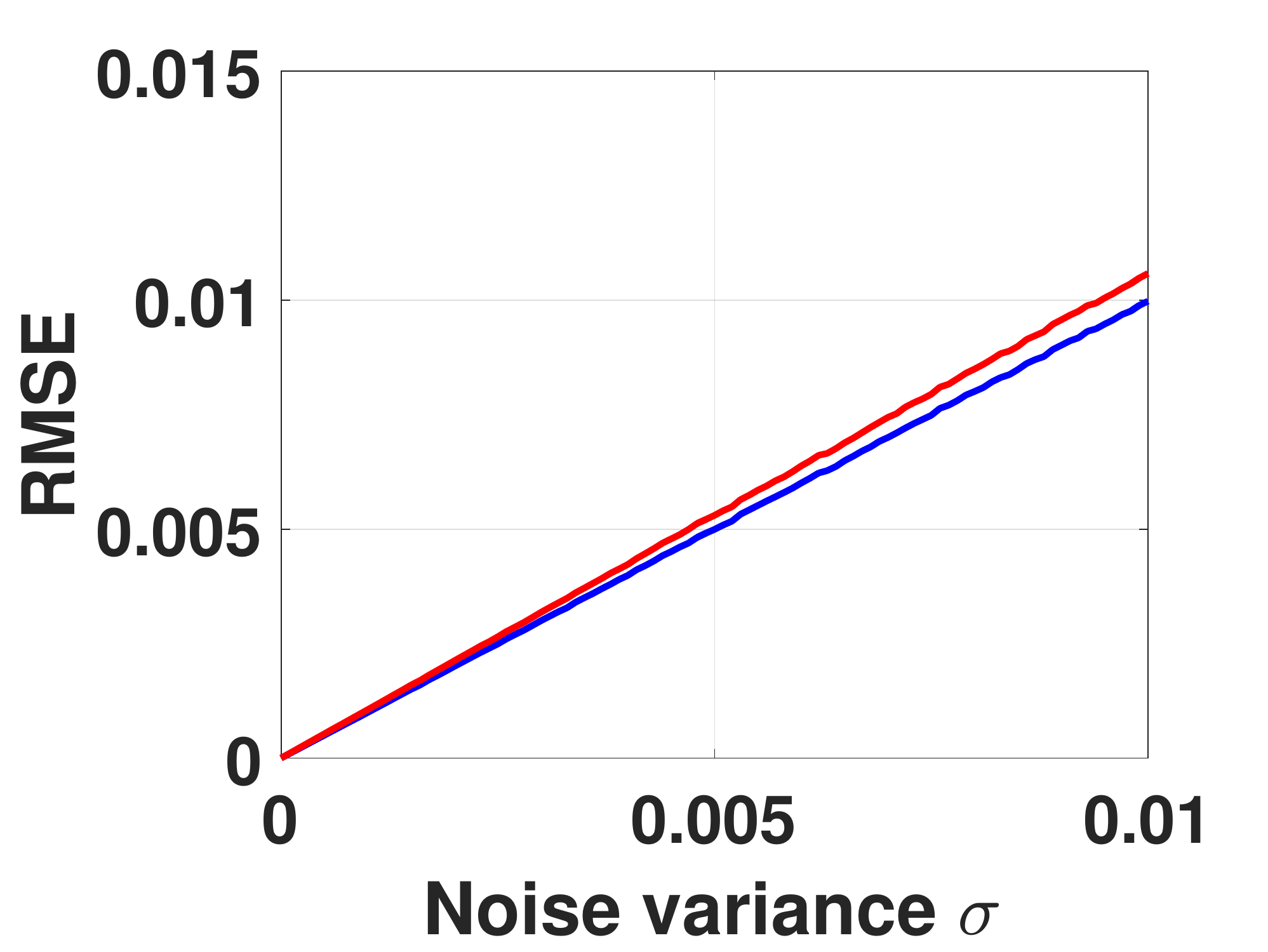}
  \label{fig:x2}
}
\subfigure[$x^3$]{
  \centering
  \includegraphics[width=0.23\columnwidth]{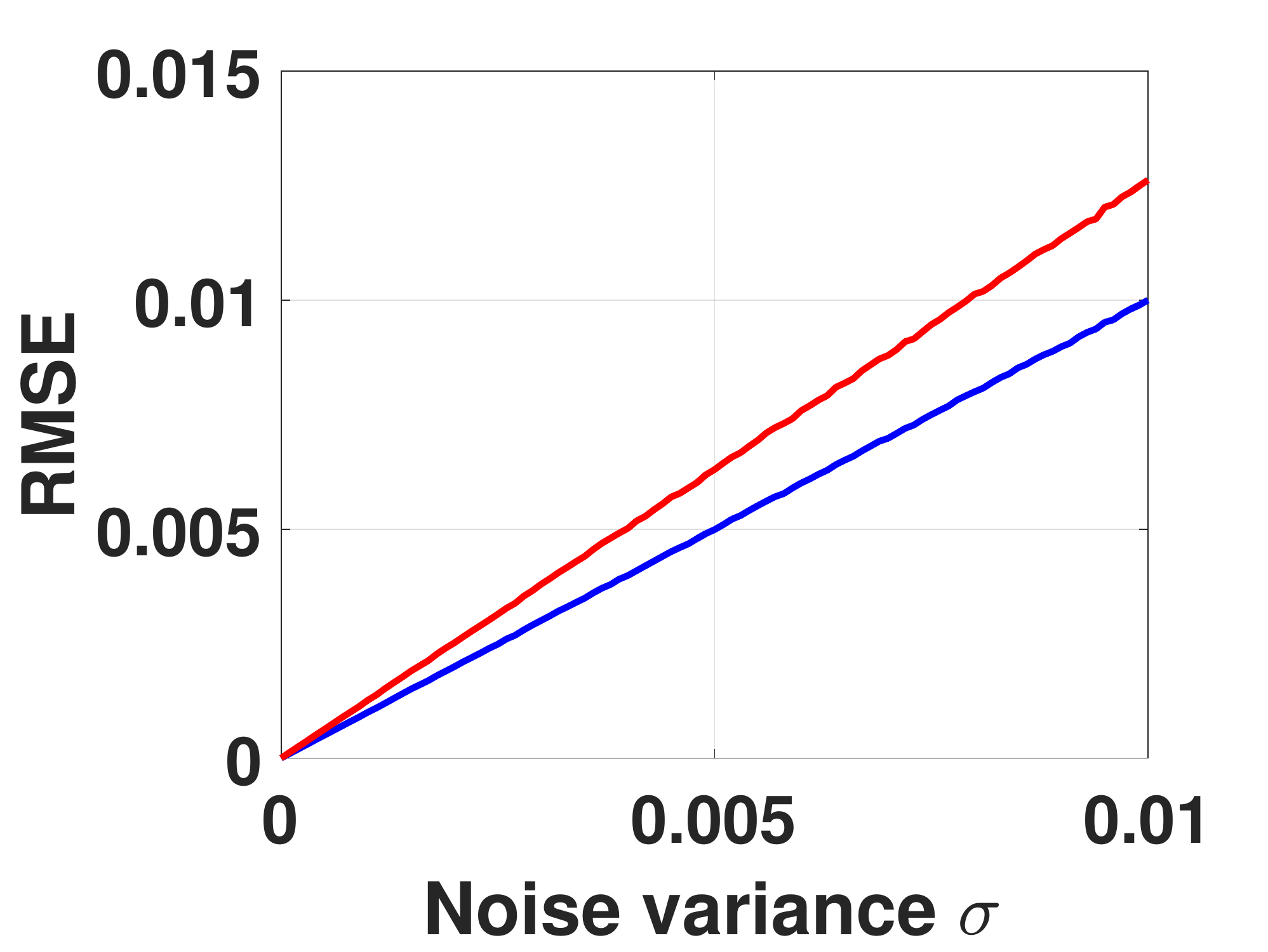}
  \label{fig:x3}
}
\subfigure[$x^5$]{
  \centering
  \includegraphics[width=0.23\columnwidth]{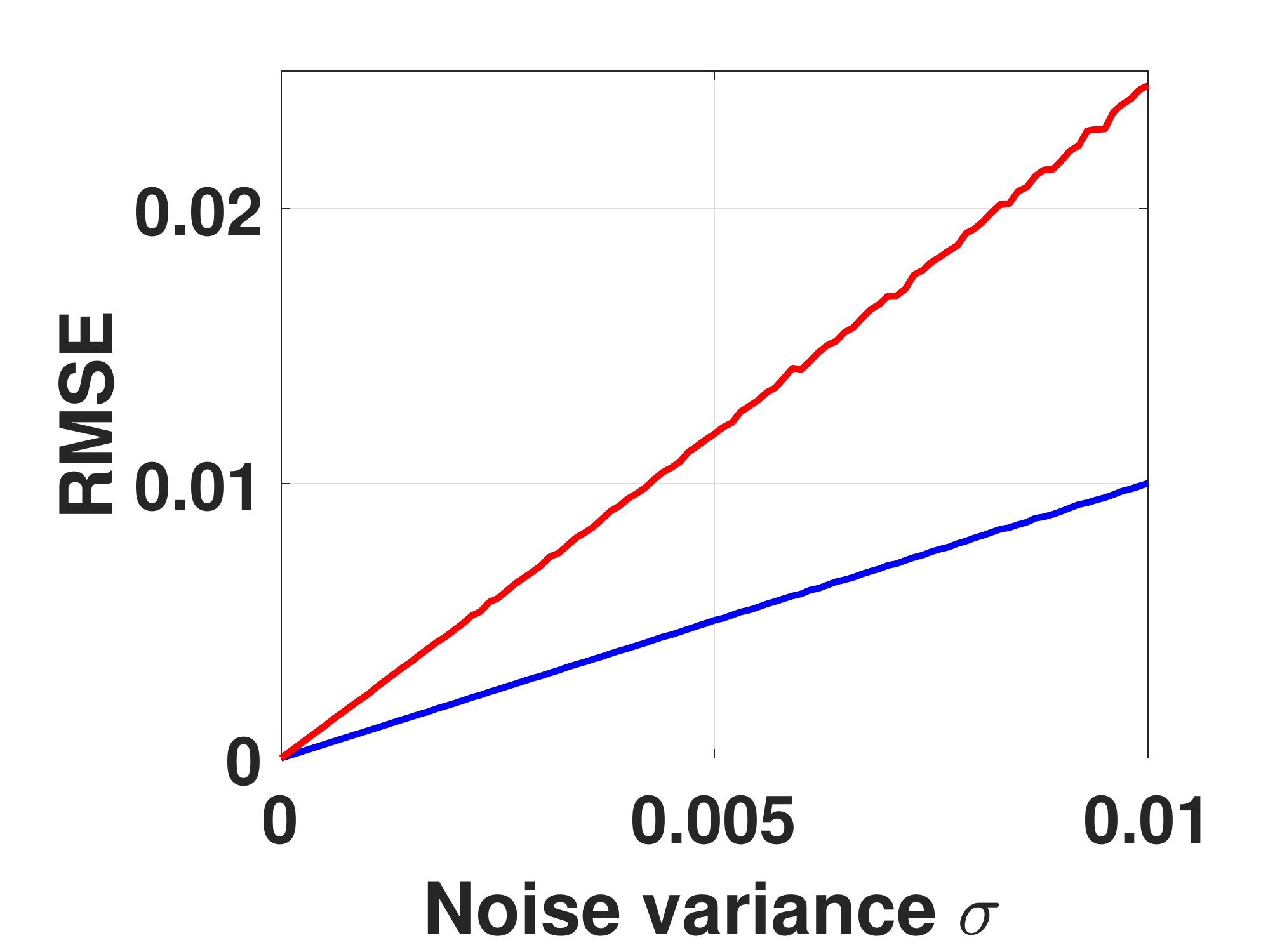}
  \label{fig:x15}
}
\subfigure{
  \centering
  \includegraphics[width=0.24\columnwidth]{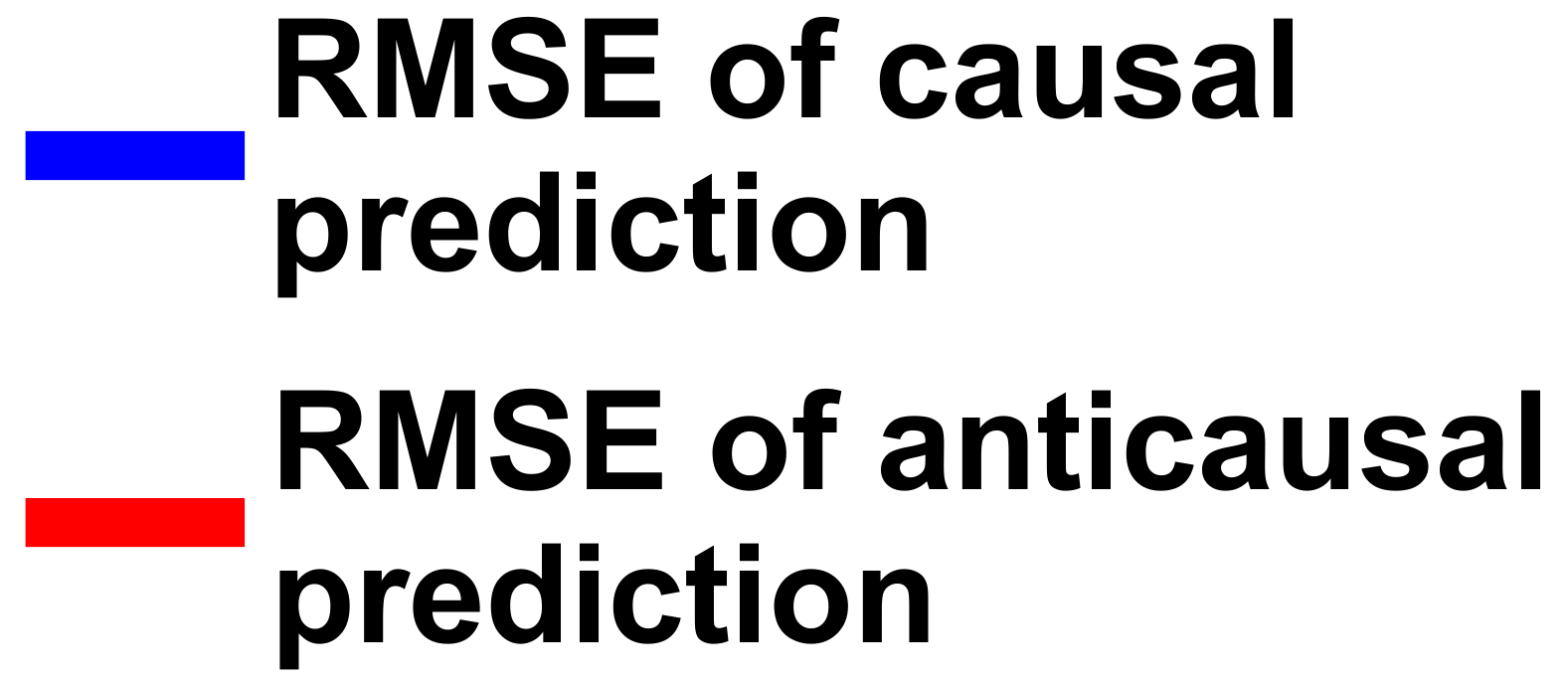}
}
\caption{Overview of the RMSE of predicting in causal and anticausal direction in all artificial data sets. The figure captions denote the corresponding function $f$. In all cases, the RMSE of causal prediction is smaller or equal to the RMSE of anticausal prediction. The difference becomes bigger with a higher variance in the noise and a higher degree of non-linearity of $f$. Note that due to the different nature of the functions, the values, and thus the axes, are differently scaled.}
\end{figure}

In this first setting, we assume to know the oracle function $\phi$ and the true causal direction. Here, we want to evaluate if the theorem holds in an optimal setting. For every function, we tested different values of the noise variance $\sigma$ with each 1000 data samples and averaged over 100 runs. According to Theorem \ref{eq:theorem}, the RMSE difference should increase with an increase in the noise variance $\sigma$ and in the degree of non-linearity of $\phi$. In the linear case, the RMSE of causal and anticausal prediction should be equal as a direct consequence of \eqref{eq:antiFinal}. Table \ref{tabel:linear} shows the results of the linear function $\phi(x) = x$ which support this conclusion. The RMSE of predicting in causal and anticausal direction are the same and perfectly represent the variance of the noise.

In case of the exponential function $f(x) = \operatorname{exp}(a \cdot x)$, the inverse is given by $f^{-1}(y) = \frac{\operatorname{log}(y)}{a}$. The results of different values for $a$ with an increasing noise variance are summarized in Figures \ref{fig:exp11} to \subref{fig:exp15}. Also these results conform with the theorem. The RMSE in causal direction always approximately represents the noise variance, while the RMSE in anticausal direction is always bigger than in causal direction when the noise variance is greater than 0.

Figure \ref{fig:sinnd} shows the result for $f(x) = \operatorname{sin}\left(\frac{(30 \cdot x + 25) \cdot 2\pi}{360}\right)$ with the inverse $f^{-1}(y) = \frac{asin(y) \cdot 360}{2 \cdot \pi} - 25$. Here, the RMSE difference also increases with an increased noise variance, but less extreme as e.g. in the case of $\phi(x) = \operatorname{exp}(5 \cdot x)$ in Figure \ref{fig:exp15}. Seeing Figure \ref{fig:Slope} and equation \eqref{eq:antiFinal}, this can be explained by the higher slope of the exponential function than that of the sinus function. In case of the exponential function, most data points fall in regions with a small slope of $\phi$, while in case of the sinus function, only slightly more data points fall in regions with a small slope than in regions with a high slope.

The last function is the power function $f(x) = x^a$, where the inverse is defined by $f^{-1}(y) = y^{\frac{1}{a}}$. Figures \ref{fig:x02} to \subref{fig:x15} show the results for various values of $a$. Again, these results conform with the theorem and are similar to the results of the exponential function.

The theorem holds in all data sets. The prediction error in causal direction was always smaller or equal to the prediction error in anticausal direction. As expected, the magnitude of the error difference greatly depends on the noise variance and on the degree of non-linearity of $\phi$. The higher the non-linearity and/or the noise variance, the higher the difference.

\subsubsection{$\phi$ and causal direction are unknown}
\label{ref:arti2}
\begin{figure}[t]
\subfigure[$x$]{
  \centering
  \includegraphics[width=0.23\columnwidth]{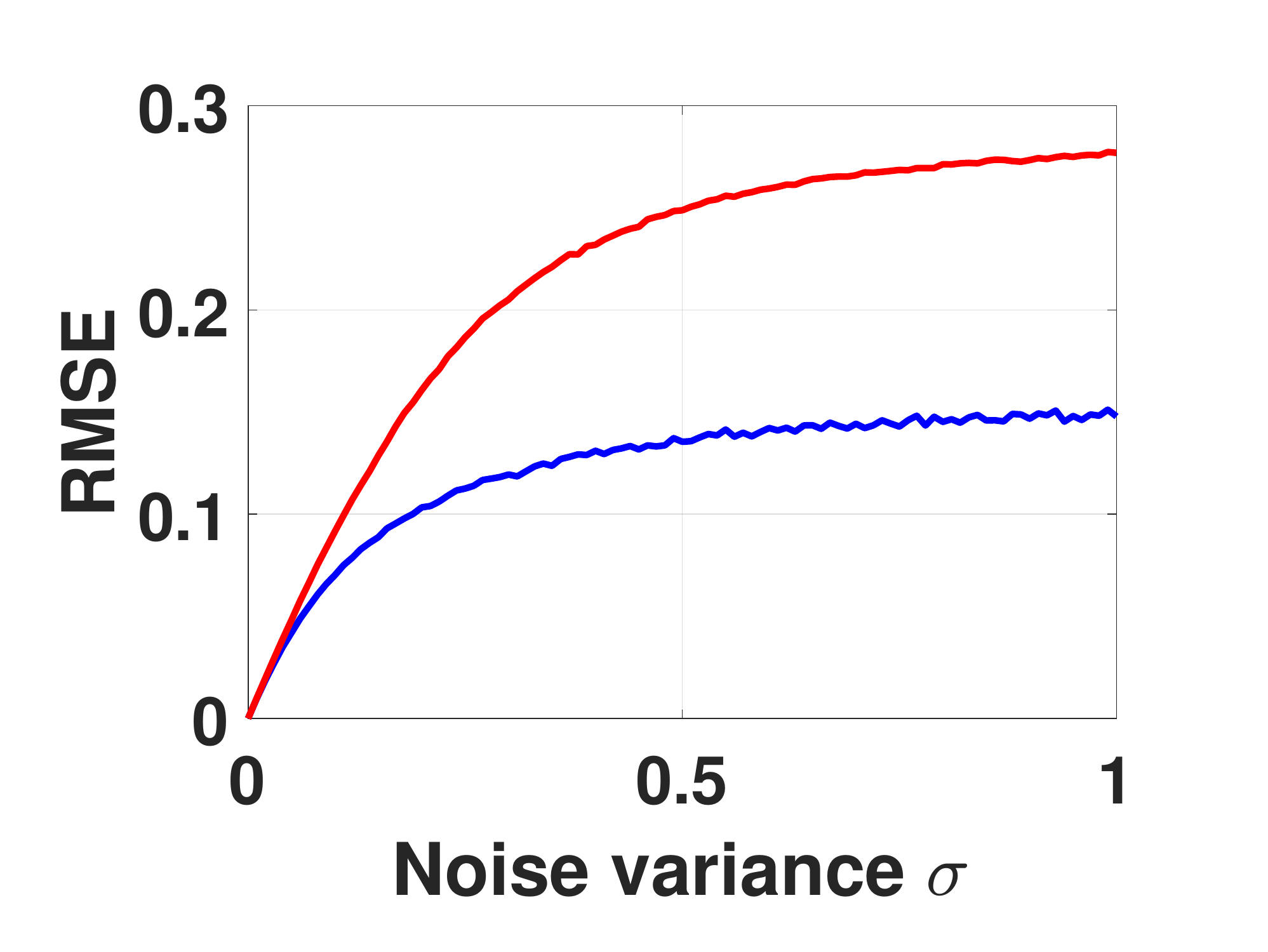}
  \label{fig:smoothingx}
}
\subfigure[$x^5$]{
  \centering
  \includegraphics[width=0.23\columnwidth]{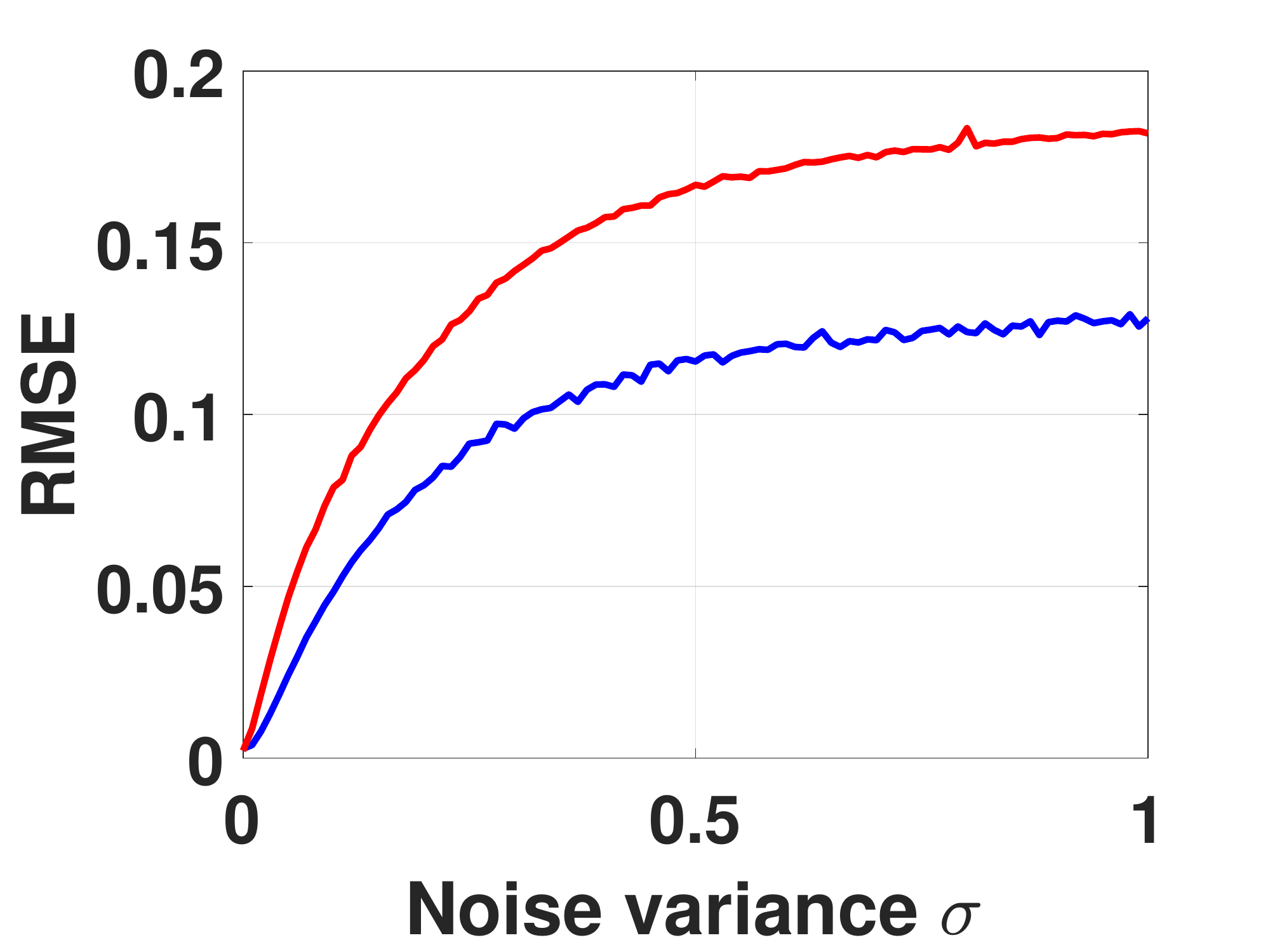}
  \label{fig:smoothingx5}
}
\subfigure[$\operatorname{exp}(x)$]{
  \centering
  \includegraphics[width=0.23\columnwidth]{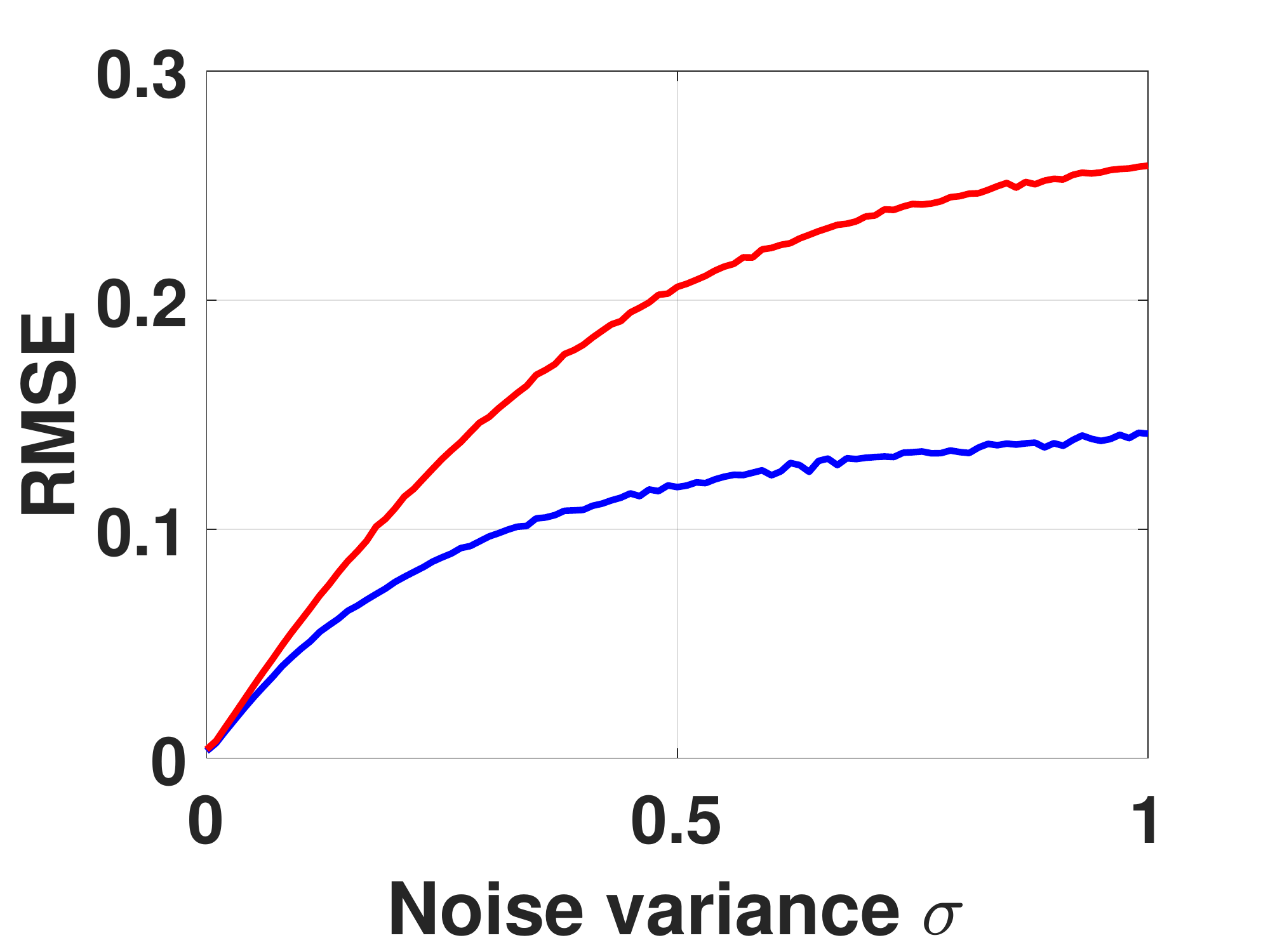}
  \label{fig:smoothingexp}
}
\subfigure{
  \centering
  \includegraphics[width=0.24\columnwidth]{Legend}
}
\addtocounter{subfigure}{-1}
\subfigure[Causal least-squares solution of $\operatorname{exp}(x)$]{
  \centering
  \includegraphics[width=0.48\columnwidth]{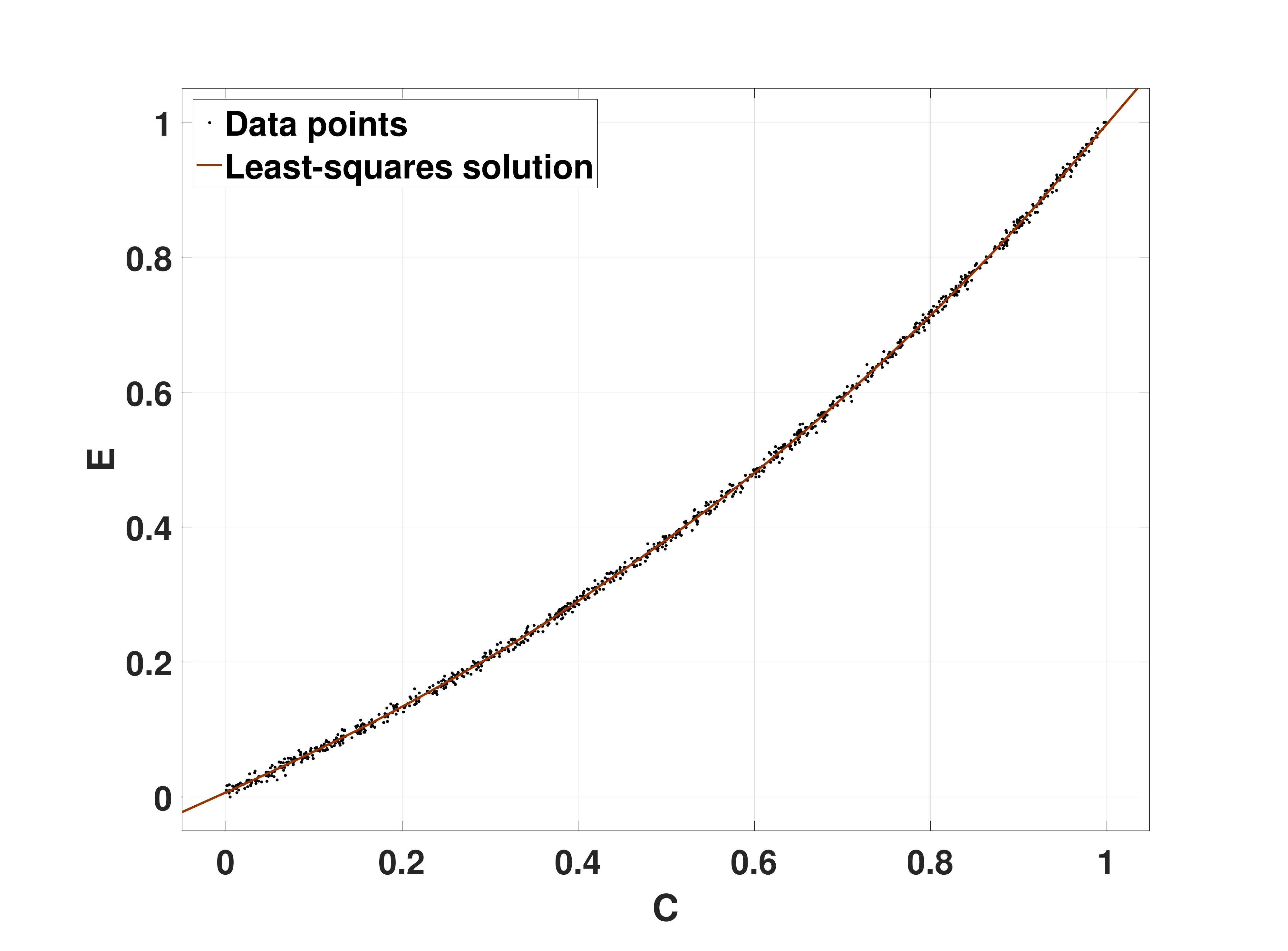}
  \label{fig:lsqcausal}
}
\subfigure[Anticausal least-squares solution of $\operatorname{exp}(x)$]{
  \centering
  \includegraphics[width=0.48\columnwidth]{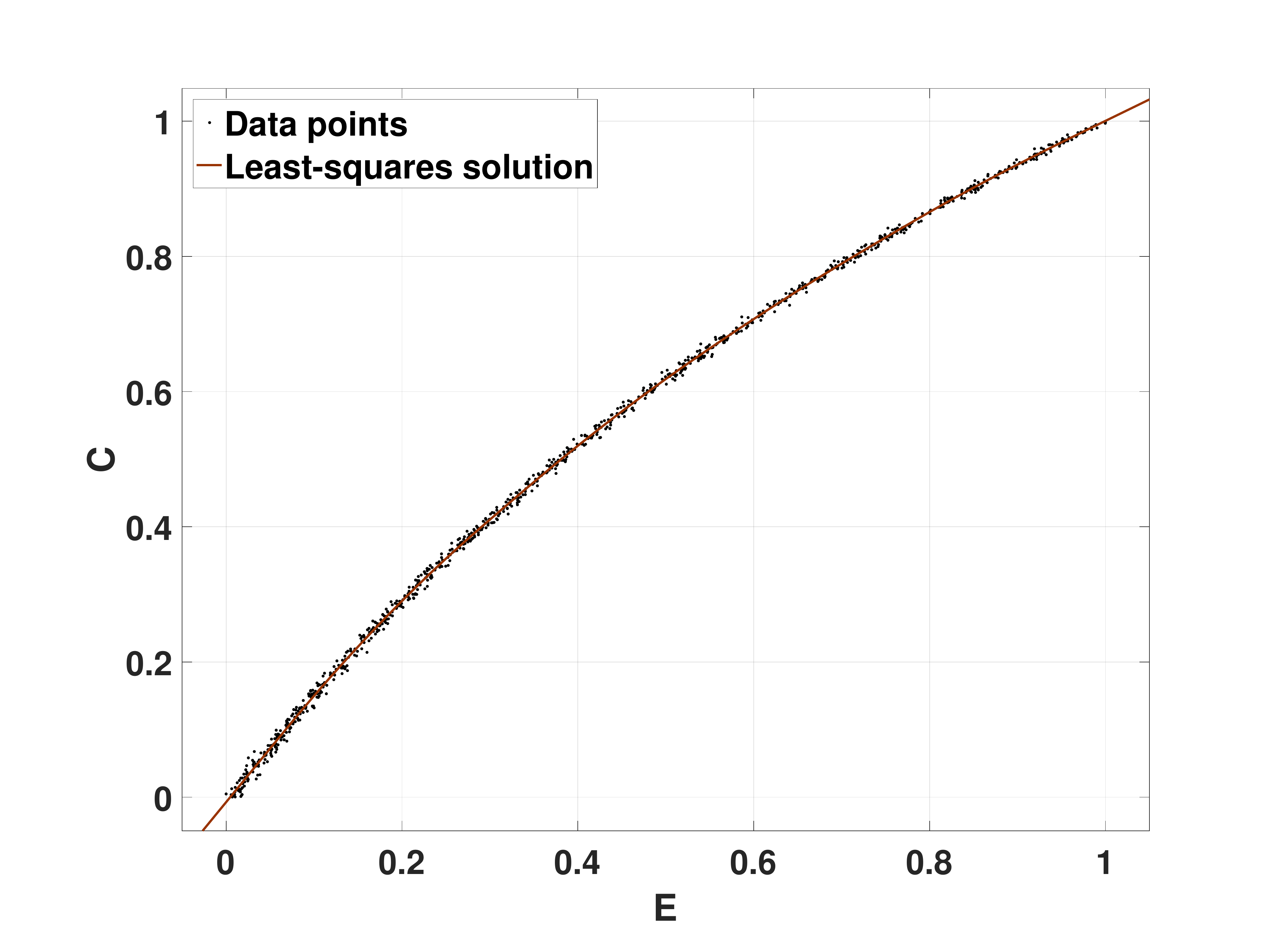}
  \label{fig:lsqanticausal}
}
\caption{Overview of the RMSE using the least-squares solutions in causal and anticausal direction for the prediction. The figure captions denote the corresponding function $\phi$. In all cases, the RMSE of predicting in causal direction is smaller than or equal to the RMSE of predicting in anticausal direction. The difference becomes bigger with a higher variance in the noise. As an example, the Figures \textbf{(d)} and \textbf{(e)} show the least-squares solutions in causal and anticausal direction for $\sigma = 0.01$, respectively.}
\end{figure}

In this setting, we assume $\phi$ and the true causal direction are unknown. These evaluations investigate how robust the theorem is with respect to merely obtaining an estimation of $\phi$ and $\phi^{-1}$, which might be a more realistic scenario. For the evaluation, we generate data similar to the aforementioned data generation process, but, since we assume to not know the causal direction, we normalized the whole effect data between $0$ and $1$ instead of only the functional output. As regression model we utilized a smoothing spline that was fitted via least-squares error minimization in both causal directions. A smoothing spline is a very flexible regression model and is less sensitive to the choice of parameters than e.g. Gaussian Process Regression. The noise variance is varied in the range $\sigma \in [0, 1]$.

However, we want to clarify that our theorem makes no statement about the expected error of the least-squares solution in anticausal direction, but it does about the expected error of $\phi^{-1}$. According to \eqref{eq:limInverse}, this experiment is justified by the idea that the least-squares solution approximately represents $\phi^{-1}$ in anticausal direction if the noise variance is sufficiently small. The cases with big noise variance are not covered by our theorem and rather serve for gaining additional insights towards an error asymmetry with respect to the least-squares solutions. Further, these experiments may indicate how the theorem can be applied to a practical domain, such as causal inference.

Figure \ref{fig:smoothingx} to \ref{fig:smoothingexp} summarize the RMSE for the functions $\phi(x) = x$, $\phi(x) = x^5$ and $\phi(x) = \operatorname{exp}(x)$. According to Corollary \ref{eq:normalizedData}, the RMSE in causal regression should be smaller than the RMSE in anticausal regression if $\operatorname{Var}[N_E] > 0$. The experimental results capture this very well even though our theorem does not provide any statements for highly noisy data. The Figures \ref{fig:lsqcausal} to \ref{fig:lsqanticausal} show two examples of the least-squares solution in causal and anticausal direction for $\sigma = 0.01$, where the RMSE of the causal solution is around $0.0057$ and the RMSE of the anticausal solution is around $0.0064$. The difference in this case might not be big, but the evaluations show a systematic asymmetry.

\subsection{Real-world data sets}
For the evaluations with real-world data sets, we used the \emph{CauseEffect} \cite{causeEffect} real-world benchmark data sets for causal inference.\footnote{Parts of the data sets are from \cite{Lichman:2013}.} The data set is a collection of various cause and effect pairs that provide knowledge about the true causal relationship between two variables. A further description can be found in \cite{Mooijetal16}.

In real-world data, the true $\phi$ is generally unknown, therefore we assumed $\phi$ to be the power function $\phi(x) = a \cdot x^b$. The power function provides a monotonic increasing behavior, offers a direct interpretation of the parameter $b$ as a measure of non-linearity and is more flexible than e.g. the exponential function. We used 92 data sets in total, where we omitted data sets with multiple variables and 3 data sets with an extreme performance gap between causal and anticausal prediction.\footnote{In these three data sets the RMSE of causal prediction was much smaller than the RMSE of anticausal prediction.} Further, if the data is approximately monotonically decreasing, the sign of the effect data was changed $E := -E$. Similar as in the second experiment with the artificial data sets, we normalized the cause and effect data between $0$ and $1$.

Based on the normalized data, the parameters $a$ and $b$ of the power functions were estimated in causal direction. These parameters were also used for the inverse model $\phi^{-1}$, which may differ from the optimal least-squares solution. Note that since the function only roughly approximates the true oracle function, the normalization is inaccurate, and thus, the parameter $a$ also needs to be estimated. As in the evaluations with artificial data sets, we then compared the RMSE of predicting in causal and anticausal direction using the estimated power function. In 88 of 92 (\textbf{95.65\%}) data sets, the RMSE of causal prediction was smaller than that of anticausal prediction. An overview of all results are given in Figure \ref{fig:real-world}, where the parameter $b$ of the model indicates the degree non-linearity. The error difference in cases for $b > 1$ is marginally in favor for anticausal regression and increases very slightly with an increase in $b$, which can be explained by small fluctuations from the noise or inaccurate model estimations. A more accurate estimation of the true $\phi$ may reveal a more clear performance gap.

Overall, surprisingly, the results substantially support the theorem with respect to the observation that the data may not always fulfill the additive noise assumption. Further, cause and effect probably have many unobserved common causes, which seem to be irrelevant for the theorem. This indicates some robustness to a violation of the made assumptions. Note that an evaluation regarding the comparison of the RMSE of the least-square solution in these real-world data could be interesting for a possible application to causal inference, but is not justified by the theorem due to the uncontrollable experiment setting and violations of the made assumptions. A further analysis regarding causal inference might be interesting for future work, but is beyond the scope of this paper.

\begin{figure}[t]
\centering
  \includegraphics[width=1\columnwidth]{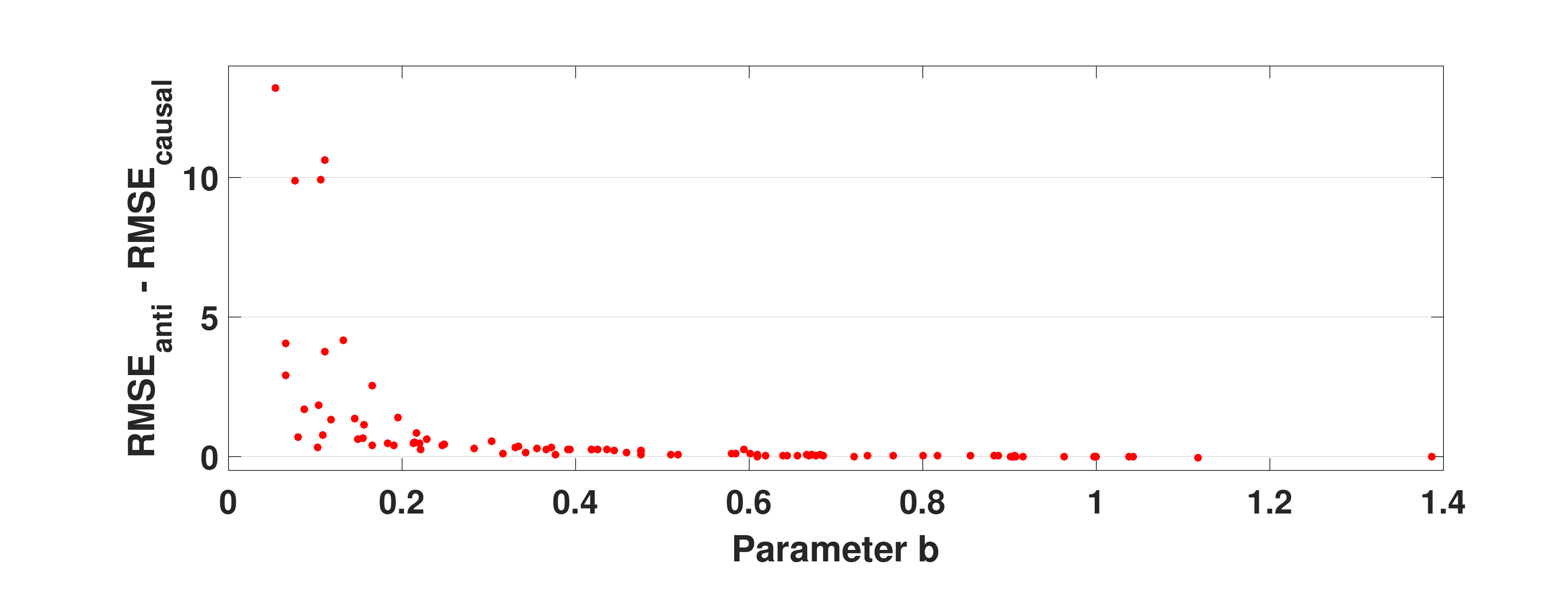}
\caption{A plot of the RMSE differences in all 92 data sets with respect to the parameter b of the prediction model.}
\label{fig:real-world}
\end{figure}

\section{Related Work}
\label{sec:related}
The expected prediction errors of various machine learning methods have been extensively studied in the past. However, the classical learning theories, such as \cite{vapnik1982estimation,bartlett2006convexity,bousquet2002stability,shawe1998structural,meir2003generalization}, are mainly focused on the general nature of statistical algorithms or data dependent properties, but do not consider a connection between the causality in the generation process of data and the corresponding implications for the prediction error. In terms of considering the causal nature of a problem, \cite{storkey2009training} points out that the causal direction in data matters for transfer learning tasks where the goal is to transfer knowledge from one data set to another similar data set. A more related work with respect to the independence assumption is done by \cite{JMLR:v16:janzing15a} who brought the asymmetry of the independence assumption explicitly to the context of semi-supervised learning. The argumentation is that additional input data sampled from $p(C)$ do not increase the prediction performance in causal problems, because $p(C)$ has no information about the mechanism $p(E|C)$. On the contrary, additional samples can improve the performance in anticausal data seeing that $p(E)$ is the effect in this case and therefore has information about the mechanism. This asymmetry is also exploited in terms of causal inference \cite{janzing2012information}, where the idea is to infer the true causal direction between two variables $X$ and $Y$ by comparing which variable has a higher correlation with $\phi'$. The drawback of this method is that the theoretical analysis only allows a deterministic relationship between cause and effect without noise.

In our work, we particularly point out an asymmetric relationship between the expected prediction error and the causal prediction direction while specifically allowing noise. To the best of our knowledge, this has not been explored previously.

\section{Contributions}
\label{sec:contributions}
In this section, we briefly discuss some exemplary scenarios where Theorem \ref{eq:theorem} gives direct contributions.

\textbf{Inverse vs reverse regression:} As already mentioned in Section \ref{sec:optimalDesignInverse}, the direct least squares estimator in anticausal direction (reverse regression) has a higher bias than the inverted least squares solution in causal direction. Early work comparing inverse and reverse regression argue that reverse regression should be preferred since it minimizes the squared prediction error \cite{krutchkoff1969classical}. However, this conclusion was criticized by many researchers \cite{berkson1969estimation,halperin1970inverse,parker2010prediction} who argued that the conclusions made based on reverse regression do not reflect the actual relationship between cause and effect, which is obvious regarding \eqref{eq:notEqual}. 

So far, a statement about the boundary of the expected error for the inverse regression was not clear. Therefore, Theorem \eqref{eq:theorem} gives a direct contribution to this discussion, since it provides a statement about the lower boundary on the prediction error of $\phi^{-1}$ with respect to $\phi$. Further, a violation of the theorem is probably an indication for a highly biased estimator.

\textbf{Calibration models:} Most calibration models, as introduced in Section \ref{sec:optimalDesignInverse}, suggest to invert the prediction model in causal direction for inverse predictions to tackle the problem of biased predictions. Theorem \eqref{eq:theorem} provides a general relationship between the prediction errors of the causal model and its inverse in the domain of calibration models. To the extent of our knowledge, this general relationship has not been provided by previous work in this domain without further assumptions about the model.

In terms of future work for calibration models, the theorem might be useful for developing new approaches for estimating rather complex models of $\phi^{-1}$ that can not be easily obtained by inverting $\phi$. Seeing that $\phi$ can be obtained by the least squares solution in causal direction, the theorem provides further constraints such as $\int_0^1 (\phi^{-1}{'}(\phi(C)))^2 p(C) dC \geq 1$ and $\int_0^1 \phi'(C) p(C) dC = \phi(1) - \phi(0)$ that may allow a more accurate estimation of $\phi^{-1}$ than without these constraints.

\textbf{Causal inference:} The theorem may also be exploited for a new causal inference principle based on a comparison of the prediction errors. A direct usage could be the scenario where data $X$, $Y$ are observed and the function $f: X \rightarrow Y$ is known beforehand by expert knowledge, but not the causal direction. The problem setting would be to determine whether $f = \phi$ or $f = \phi^{-1}$. This kind of scenario could e.g. be interesting for protein interactions, where the functional relationship might be known but not the causal direction. The function $f$ can be inverted and then Theorem \eqref{eq:theorem} provides a direct way to determine cause and effect based on the prediction error.

Also, as the experiments in \ref{ref:arti2} may indicate, a direct application of the theorem to causal inference might already be possible by a simple comparison of the RMSE of the least-squares solution in causal and anticausal direction under the assumption of a small error noise.

Further, the positive dependency between the error and the slope of $\phi$ in anticausal but not in causal direction could also be exploited. This may particularly be interesting for causal inference methods that test for an independence between input and prediction error \cite{hoyer2009nonlinear,zhang2009identifiability}. A comparison between error and slope may offer a more general setting regarding the assumptions about $C$ and $N_E$.

\section{Conclusion}
\label{sec:conclusion}
In this paper, we addressed the implications for the optimal prediction in causal and anticausal regression problems and explicitly allow noise affecting the effect. Under the independence assumption and additive noise model, the intrinsic causal structure can give crucial information about the prediction capabilities not only of statistical learning models but also of general prediction tasks. This has not been recognized in the past. Based on the theoretical analysis, we concluded a theorem which states that the expected prediction error of causal regression problems is smaller or equal to that of anticausal problems with respect to the normalized oracle function $\phi$ and it's inverse $\phi^{-1}$:
\begin{equation*}
	\mathbb{E}[(\phi(C) - E)^2] \leq \mathbb{E}[(\phi^{-1}(E) - C)^2]
\end{equation*}
This implies a dependency between error and slope of $\phi$ in anticausal but not in causal direction. We further empirically evaluated this theorem in various artificial and real-world data sets which give supporting results and indicate that the made assumptions are quite robust to violations. This result can significantly contribute to further theoretical and practical work.

In our future work, we further explore this error asymmetry in order to provide a new principle for causal inference. For this, we also work on a generalization of the result that allows a dependency between input variable and noise variable to make it applicable to a larger range of problems. Another interesting aspect could be the error in causal and anticausal classification problems, which may have different key factors than regression problems. An extension to the multivariate case is also of interest for future work, where an extension for $\phi: \mathbb{R}^n \rightarrow \mathbb{R}^n$ is straightforward. In case of $\phi: \mathbb{R}^n \rightarrow \mathbb{R}$, sliced inverse regression could be considered for anticausal predictions.

\section*{Acknowledgement}
This work was supported by JSPS KAKENHI \#25240036 and \#24700275 and the Center of
Innovation Program from the Japan Science and Technology Agency, JST. The final publication is available at 
Springer via http://dx.doi.org/10.1007/s41237-017-0022-z.

\bibliographystyle{unsrt} 
\bibliography{Literature}
\end{document}